\documentclass[12pt]{article}
\usepackage{graphicx}
\usepackage{float}
\usepackage{multirow}
\usepackage{longtable}
\usepackage{array}
\usepackage[table]{xcolor}
\usepackage{csquotes}

\usepackage{fullpage}
\usepackage{times}
\usepackage{xurl}
\usepackage[hidelinks]{hyperref}
\usepackage{enumitem}
\usepackage{booktabs}
\usepackage{amsmath}
\usepackage{amssymb}

\usepackage{caption}
\usepackage{newfloat}
\usepackage{placeins}
\DeclareFloatingEnvironment[name={Extended Data Fig.},fileext=ledf]{extendeddatafigure}
\DeclareFloatingEnvironment[name={Extended Data Table},fileext=ledt]{extendeddatatable}
\DeclareCaptionLabelSeparator{bar}{\enspace|\enspace}
\setlist[itemize,enumerate]{nosep,leftmargin=*}
\setlist[description]{nosep,leftmargin=!,widest={Channels 355--454}}

\title{Scalable Decision Making for Games of Imperfect Information }
\author{Samuel~Sokota\textsuperscript{1,*}, Eugene~Vinitsky\textsuperscript{2},
Hengyuan~Hu\textsuperscript{3}, Zhiyuan~Fan\textsuperscript{4},\\
J.~Zico~Kolter\textsuperscript{1}, and Gabriele~Farina\textsuperscript{4,*}\\[0.75em]
{\small\textsuperscript{1}Carnegie Mellon University, Pittsburgh, PA, USA}\\
{\small\textsuperscript{2}New York University, Brooklyn, NY, USA}\\
{\small\textsuperscript{3}Stanford University, Stanford, CA, USA}\\
{\small\textsuperscript{4}Massachusetts Institute of Technology, Cambridge, MA, USA}\\
{\small\textsuperscript{*}Corresponding authors}}
\date{}

\newcolumntype{L}[1]{>{\raggedright\arraybackslash}p{#1}}
\makeatletter
\newcommand{\naturetblrcaption}{}
\newcommand{\naturetblrlabel}{}
\newcommand{\naturetblrcolumns}{}
\define@key{naturetblr}{caption}{\def\naturetblrcaption{#1}}
\define@key{naturetblr}{label}{\def\naturetblrlabel{#1}}
\define@key{naturetblr}{columns}{\def\naturetblrcolumns{#1}}
\newcommand{\naturetblrsetup}[1]{\def\naturetblrcaption{}\def\naturetblrlabel{}\def\naturetblrcolumns{}\setkeys{naturetblr}{#1}}
\newcommand{\naturelongtblrcols}[1]{\def\naturetblrcols{@{}L{0.35\linewidth}L{0.57\linewidth}@{}}\in@{X X X X X}{#1}\ifin@\def\naturetblrcols{@{}L{0.16\linewidth}L{0.18\linewidth}L{0.18\linewidth}L{0.18\linewidth}L{0.18\linewidth}@{}}\fi
}
\newcommand{\naturetblrlabelcols}{\ifx\naturetblrcolumns\@empty\else
    \ifnum\naturetblrcolumns=5\relax
      \def\naturetblrcols{@{}L{0.16\linewidth}L{0.18\linewidth}L{0.18\linewidth}L{0.18\linewidth}L{0.18\linewidth}@{}}\fi
  \fi
}
\newenvironment{longtblr}[2][]{\naturetblrsetup{#1}\naturelongtblrcols{#2}\naturetblrlabelcols
  \begingroup
  \rowcolors{2}{gray!10}{white}\expandafter\longtable\expandafter{\naturetblrcols}\ifx\naturetblrcaption\@empty\else
    \caption{\naturetblrcaption}\ifx\naturetblrlabel\@empty\else\label{\naturetblrlabel}\fi\\\fi
  \toprule
}{\bottomrule
  \endlongtable
  \endgroup
}

\newenvironment{detailtable}[4]{\begingroup
  \rowcolors{2}{gray!10}{white}\begin{longtable}{L{0.32\linewidth}L{0.60\linewidth}}\caption{#1}\label{#2}\\\toprule
  \textbf{#3} & \textbf{#4}\\
  \midrule
  \endfirsthead
  \toprule
  \textbf{#3} & \textbf{#4}\\
  \midrule
  \endhead
}{\bottomrule
  \end{longtable}\endgroup
}

\newenvironment{channeltable}{\begingroup
  \rowcolors{2}{gray!10}{white}
  \addtocounter{table}{-1}\begin{longtable}{L{0.16\linewidth}L{0.76\linewidth}}\toprule
  \textbf{Channel(s)} & \textbf{Description}\\
  \midrule
  \endfirsthead
  \toprule
  \textbf{Channel(s)} & \textbf{Description}\\
  \midrule
  \endhead
}{\bottomrule
  \end{longtable}\endgroup
}
\makeatother

\begin{document}

\maketitle

\begin{abstract}
Real-world decision making generally involves hidden information, that is, information that is unknown to one agent but possessed by another.
Unfortunately, the presence of large amounts of hidden information renders established reinforcement learning and search approaches ineffective. Even with multimillion-dollar industrial research efforts~\cite{deepnash}, top-human-level play at Stratego---a board wargame with hidden information on a massive scale---has remained beyond the reach of AI. Here we introduce Ataraxos, an AI for Stratego based on general techniques that we developed for both self-play reinforcement learning and test-time search under hidden information. Ataraxos defeated the most decorated human Stratego player of all time by a large margin---achieving the first superhuman result in the game's history---while consuming orders of magnitude less compute and data than prior efforts. Using the same techniques, we built the first superhuman AI for Barrage Stratego and state-of-the-art AIs for Hanabi and \textit{dou dizhu}, all with low cost and high sample efficiency. The success of this approach across adversarial, cooperative, and team games establishes a design pattern for reinforcement learning and search that is effective under large amounts of hidden information, a longstanding desideratum of the field of strategic decision making.
\end{abstract}

Imperfect information~\cite{vonNeumannMorgenstern1944,kuhn1953information} is a term of art describing interactions in which some agents may possess information that others do not. It is a characterizing feature of real-world settings, including financial markets, military conflict, and negotiations.

Whereas in perfect-information settings, like chess and Go, the conceptual \textit{right thing to do} is simple (select a move that maximizes the value of the resulting position), in those of imperfect information, it is subtle. The value of a decision depends not only on the policies agents employ thereafter, but also on those they used---and counterfactually would have used---prior to and at the time of the decision (as these all affect the posterior distribution over hidden information). Even reasoning about the dependencies among contemporaneous counterfactuals in isolation is complex (see Figure~\ref{fig:dependencies}).

\begin{figure}[!htbp]
\centering
\includegraphics[width=\linewidth,height=.82\textheight,keepaspectratio]{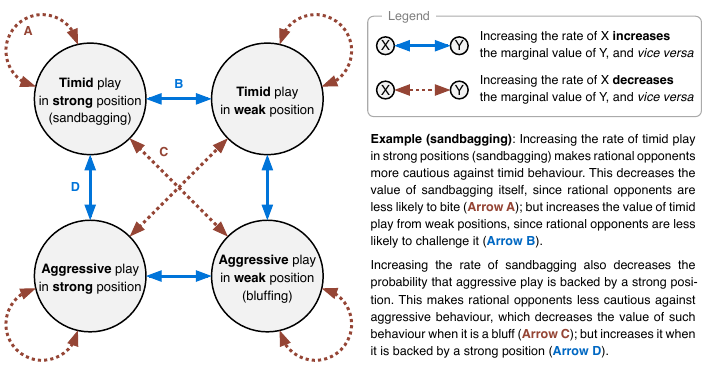}
\setcounter{figure}{0}
\caption{Graphical representation of differential relationships between decision frequencies and values among contemporaneous counterfactuals under imperfect information. Nodes represent decisions. The colour/dash pattern of an edge represents whether changing the frequency of a decision on one end of the edge increases (blue/solid) or decreases (brown/dashed) the marginal value of the decision on the other end of the edge. The interconnectedness of frequencies and values across decisions makes it difficult to intuit the frequencies with which decisions should be made.}
\label{fig:dependencies}
\end{figure}

These dependencies have made developing AI for imperfect-information settings challenging. The most successful approaches use sophisticated problem transformations based on public information~\cite{deepstack,safe_nested,libratus,modicum,pluribus,supremus,rebel,aii23}. But the cost of these transformations scales with the amount of hidden information, making them applicable only when this amount is small, as in Texas hold'em (in which there are 1,326 possible hands). Due to this fundamental limitation, and the absence of an alternative foundation, strategic decision making in settings with large amounts of hidden information has remained an open problem.

The unresolvedness of this problem is epitomized by the state of AI for Stratego---a board wargame resembling military chess (see Figure~\ref{fig:example-board}) prized as a challenge problem for the scale of its hidden information (there are over $10^{33}$ possible piece configurations). Stratego has been the subject of industrial research efforts spanning multiple years, involving dozens of researchers, and expending computation costing millions of dollars~\cite{deepnash}. Yet, human players have remained superior, potentially making it the only classical game in which such well-resourced efforts have failed to produce superhuman performance.

\begin{figure}[!htbp]
\centering
\includegraphics{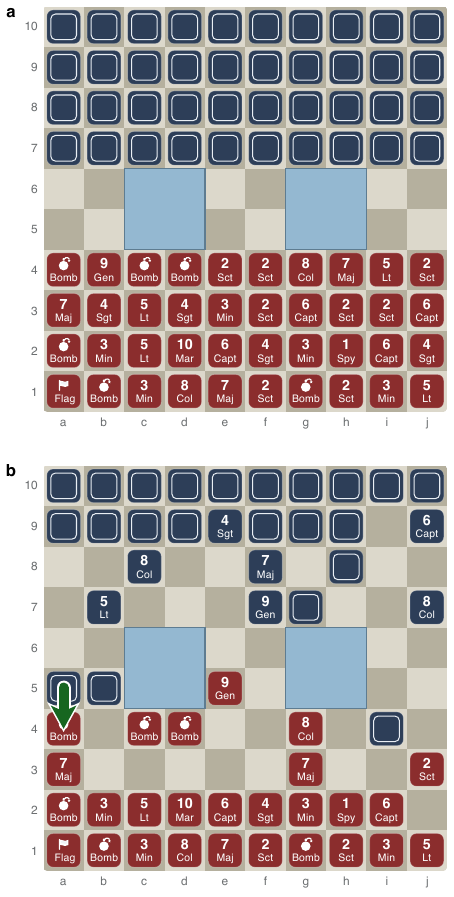}
\setcounter{figure}{1}
\caption{An illustration of the game of Stratego. \textbf{a}~The starting position of a game from the perspective of the red player. Each player arranges their pieces in secret, so the types of the blue pieces are concealed from the red player (and vice versa). From this starting position, the players alternate moves, each of which is defined by the displacement of a single piece, as in chess. \textbf{b}~A move in the game for the blue player. On this move, because a blue piece moves onto a square occupied by a red piece, there is a battle. During a battle, the types of both pieces are revealed and at least one of the two is captured (determined based on their types and attack-defence roles). If a Flag (see piece on a1) is captured in battle, the capturing player wins the game.}
\label{fig:example-board}
\end{figure}

Here, we introduce Ataraxos, an AI for Stratego.
Ataraxos was built using general techniques that we developed for self-play reinforcement learning and test-time search under massive amounts of hidden information. In a 20-game series, Ataraxos defeated Pim Niemeijer---the most decorated Stratego player of all time---by a margin of victory without precedent at the highest level of play: 15 wins, 1 loss, and 4 draws. Ataraxos achieved this result while costing only a few thousand dollars to train.

To demonstrate generality, we applied these techniques to build AIs for Barrage Stratego, a Stratego variant; Hanabi, the premier benchmark for cooperative imperfect-information games; and \textit{dou dizhu}, one of the most popular card games in China. Our AIs defeated three multi-time world champions in Barrage Stratego (the first superhuman result for the game), achieved a new state of the art for Hanabi, and beat the state-of-the-art bots for \textit{dou dizhu}.

The success of these techniques across adversarial, cooperative, and team games shows that reinforcement learning and search are no longer precluded from high performance by the presence of large amounts of hidden information. This marks the fulfilment of a longstanding aspiration of the field of strategic decision making.

\section{Overview of Ataraxos}

Ataraxos is based on \textit{tabula rasa} self-play reinforcement learning and test-time search---a design pattern that is general across imperfect-information games. The pattern comprises a policy--value network, trained through self-play to propose decisions and predict game outcomes; a belief network, trained on self-play data to model hidden information; and a search procedure that refines the policy at test time.

In self-play reinforcement learning, the policy--value network improves by playing games against itself and shifting probability towards decisions that lead to wins. The core innovation that enables effective reinforcement learning is a coordination between regularization strength and policy update size~\cite{mmd}. Ataraxos employs stronger regularization and more aggressive updates early in self-play, and weaker regularization and smaller policy updates late in self-play. Two complementary considerations motivate this interplay. First, keeping update sizes commensurate with regularization strength damps the otherwise cyclical, divergent, or chaotic learning dynamics engendered by imperfect information, leading to stable progress. Second, adapting both quantities over the course of training facilitates rapid improvements early on and continued improvement later---avoiding both update sizes so small as to make improvement impractical and regularization so large as to compromise the policy.

Ataraxos trains a belief network on the self-play games of its final policy--value network. Given the information available to the player at a position, this network is trained to model the hidden information---whose ground truth is available in self-play. This trained belief network serves as a generative model, allowing Ataraxos to sample realizations of hidden information based on their likelihoods.

At test time, Ataraxos selects its actions by search. To start, it generates realizations of hidden information with its belief network. Then, with its policy--value network, it plays out candidate actions under each realization and estimates the values of these actions from the positions reached. Finally, treating these estimates as if they had come from self-play, Ataraxos performs one additional update step---applied only to the current decision---and selects its action from the updated policy. Because this step mimics those of self-play reinforcement learning, the search inherits improvement properties thereof~\cite{ues}.

For Stratego, Ataraxos instantiates two interdependent self-play processes, corresponding to the two phases of the game: in one, a setup network learns to arrange the 40 pieces; in the other, a move network learns to move the pieces. The two processes are coupled through the games themselves---the setup process supplies the initial boards on which the move process plays, and the outcomes of the resulting games drive the updates of both. The belief network learns to predict the types of the opponent's hidden pieces. At test time, setups are sampled directly from the setup network, while moves are selected using search.

We trained the resulting Stratego system at a total cost of a few thousand dollars. This economy reflects both a fast implementation---a custom CUDA~\cite{cuda} simulator that executes millions of state updates per second---and sample-efficient learning.

\section{Evaluation} \label{sec:evaluation}

We evaluated Ataraxos against Pim Niemeijer~\cite{pim_kleier,stratego_wikipedia} (hereafter Pim), a player whose accolades include:
\begin{itemize}
  \item 4 world championships (the most of any active player and tied for the most of all time);
  \item 15 Dutch national championships (the most of all time);
  \item 2 online world championships (the most of all time);
  \item over 600 weeks as the \#1-ranked player (the most of all time).
\end{itemize}
According to George Franka~\cite{george_franka_kleier}, the only player to compete in every world championship since 1997, Pim is ``the best Stratego player ever.''

The evaluation consisted of a 20-game series, with performance measured by effective win rate (i.e., by counting draws as half wins). The large number of games was chosen both to reduce variance (because strong play in Stratego requires substantial randomization, competent players win games against top players more often than they would in games like chess) and to give Pim an opportunity to find and exploit weaknesses in the strategy of Ataraxos---which, he was informed, would not adapt to his play. To prevent fatigue and allow time for tactical preparation between games, the evaluation was spread over the course of three weeks. Pim was paid \$1,000 for participating in the evaluation, with an additional \$100 for each win and \$50 for each draw to align his monetary incentives with his performance.

Ataraxos won the series with 15 wins, 1 loss, and 4 draws (see Figure~\ref{fig:wp}). This margin (an 85\% effective win rate) is without precedent at the highest level of human play, where, according to three-time runner-up world champion Max Roelofs~\cite{max_roelofs_kleier}, margins are razor thin due to the risk that must inevitably be assumed during play. Ataraxos achieved such a margin despite the asymmetric structure of the evaluation---wherein Pim could adapt to Ataraxos over a large number of games, but Ataraxos could not adapt to Pim---which, according to three-time world champion Vincent de Boer~\cite{vincent_de_boer_kleier}, constituted a large handicap.

\begin{figure}[!htbp]
\centering
\includegraphics[width=\linewidth,height=.82\textheight,keepaspectratio]{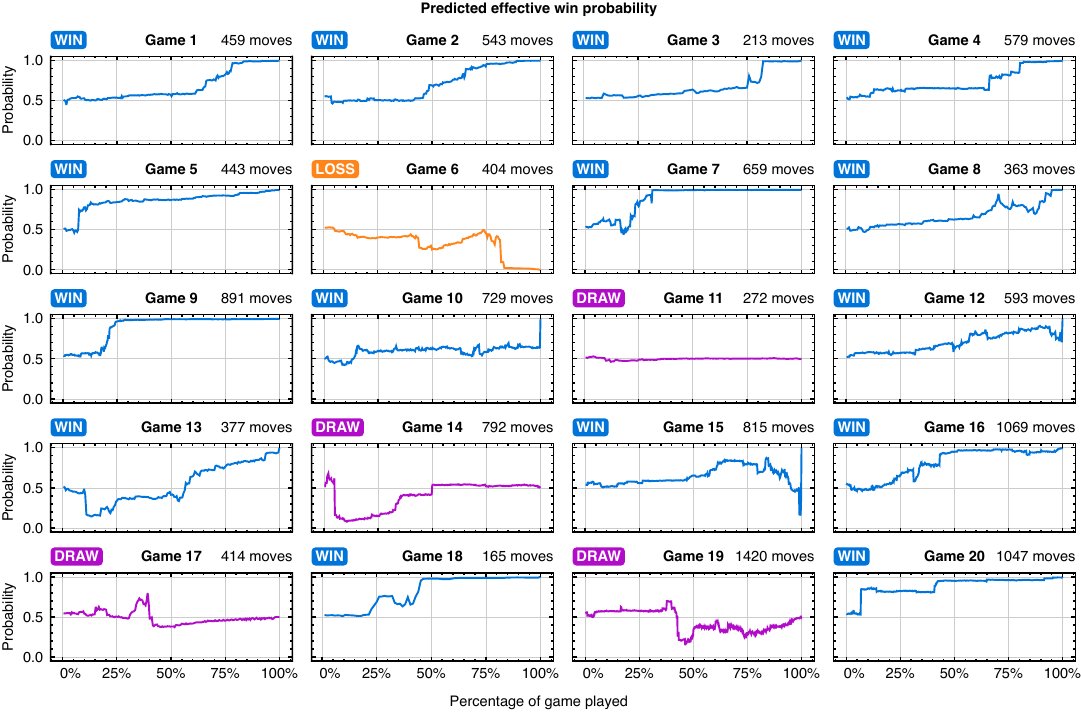}
\setcounter{figure}{2}
\caption{Effective win probability implied by the network value predictions over the games played against Pim Niemeijer. Effective win probability is the expected value of the game with winning assigned value 1, drawing assigned value 1/2, and losing assigned value 0.}
\label{fig:wp}
\end{figure}

Because of this adversarial adaptation, and more broadly because human strategies are not static from game to game, the outcomes of the games were far from independently and identically distributed. But under the assumption that they had been, an exact one-sided binomial test of whether Ataraxos was more likely to win than lose would have yielded a $p$-value less than~$2.6 \times 10^{-4}$.

Following the evaluation against Pim, we demoed Ataraxos at the 2025 Stratego World Championship from August 1 to 3. During the demo, world championship attendees were given the opportunity to play against Ataraxos. Across 40 such games, Ataraxos recorded a 95\% effective win rate (38 wins, 2 losses, 0 draws), providing additional evidence of strength against a range of players and playing styles.

\section{Application to Other Games}

We applied the techniques underlying Ataraxos to build AIs for Barrage Stratego, Hanabi, and \textit{dou dizhu} (each of which we also refer to as Ataraxos, after their shared design), thus spanning two-player zero-sum, fully cooperative, and two-team zero-sum games. In all cases, we achieved state-of-the-art results, illustrating the effectiveness of these techniques across a diverse collection of imperfect-information settings.

\subsection{Barrage Stratego}
Barrage Stratego is a competitively played variant of Stratego involving 8 pieces per player (as opposed to the classic variant, which involves 40). Games of Barrage are more precarious than those of the classic variant, and good strategy requires a higher degree of risk taking, bluffing, and sandbagging.

We evaluated our AI for Barrage Stratego against three of the four top-ranked human players, each of whom is a two-time Barrage World Champion. The evaluations consisted of four 50-game series (one each against the \#3- and \#4-ranked players and two against the \#1-ranked player). We used a large number of games both because games of Barrage are typically shorter than those of classic Stratego and because individual game outcomes tend to be more stochastic than in the classic variant.

Ataraxos won each of the four series. With the same caveats and under the same assumptions as in the \nameref{sec:evaluation} section, an exact one-sided binomial test of its aggregate record would have yielded a $p$-value less than $1.3 \times 10^{-5}$.
These results are the first in which AIs have outperformed top humans at Barrage Stratego.

\subsection{Hanabi}

Hanabi is the most popular benchmark among fully cooperative imperfect-information games~\cite{hanabi-challenge} and the winner of the 2013 Spiel des Jahres award for best board game of the year. In the game, players jointly build five colour-coded fireworks stacks in ascending order, each player holding their cards facing outward so that only teammates can see them. On each turn, a player either reveals limited information about the cards of a teammate, discards a card, or attempts to extend one of the stacks, with the final score determined by the number of cards successfully played. The number of deals exceeds $5 \times 10^{27}$ for the five-player variant.

We implemented Ataraxos for each of the two- to five-player variants of the game.
Ataraxos set a new state of the art with statistical significance for all variants.
For the two-player variant, our result required two orders of magnitude less compute than the previous state of the art.

\subsection{\textit{Dou Dizhu}}

\textit{Dou dizhu} is one of the most popular card games in China, with millions of active players and mobile apps amassing billions of downloads \cite{ContributorstoWikimediaprojects2026Feb,doudizhupop}.
One player, the landlord, competes against a team of two peasants in a three-player, asymmetric partnership game. The landlord begins with a larger hand and plays alone, while the two peasants cooperate implicitly through their actions. The objective is to be the first to empty one's hand by playing legal card combinations of progressively higher rank, with players alternating turns and either beating the current combination or passing. The number of possible deals in the game exceeds $3 \times 10^{24}$.

We measured the performance of our approach against the previous state of the art, PerfectDou~\cite{yang2022perfectdou}, as well as its predecessor DouZero~\cite{zha2021douzero}. Following PerfectDou, the evaluation treated the game as a player-vs-team setting, where one agent controls the landlord and the other controls both peasants in a decentralized fashion. Ataraxos defeated PerfectDou and DouZero with statistical significance, establishing a new state of the art.

\section{Conclusion}

The combination of reinforcement learning and search has proven to be a powerful recipe for strategic decision making~\cite{tdg,ag,agz,muzero,katago,bitterlesson,diplodocus}. But in the past, this recipe has had limited applicability to imperfect-information settings. The success of the AIs presented herein shows that the presence of large amounts of hidden information is no longer prohibitive. This puts practical AI within reach for many strategic decision-making problems for which fast, accurate simulators can be constructed.

\FloatBarrier

\clearpage
\part*{Methods}

\section{Design of Ataraxos}

Ataraxos consists of two interdependent self-play reinforcement learning processes, realized by transformer networks for setup selection and move selection; a belief network trained on self-play games of the final selection networks; and a test-time search procedure that composes the move and belief networks.

\subsection{Interdependent Self-Play Processes}

The foundation of Ataraxos is its self-play reinforcement learning module. This module comprises two separate but interdependent self-play processes associated with the two phases of Stratego. The first phase, in which players privately determine starting positions for their pieces---called setups---is handled by one self-play process. The second phase, in which players alternate moving their pieces, is handled by another. These processes learn separately but in tandem: the setup selection process determines the initial boards for the move selection process; and the move selection process determines game outcomes that both processes use for policy updates. We chose this decomposition rather than an end-to-end model because, although an end-to-end approach would avoid making the phase decomposition explicit, it would require a single model and training pipeline to accommodate two unrelated tasks. As discussed in the Methods subsections \nameref{sec:architectures}, \nameref{sec:data-generation}, and \nameref{sec:damping}, with full Stratego training specifications provided in the Supplementary Information, setup learning favoured a decoder-only architecture, Monte Carlo expected return and advantage estimation, and higher learning rates and regularization temperatures. Move learning, by contrast, favoured an encoder-only architecture, $\lambda$-based expected return and advantage estimators, an annealed learning-rate schedule, and lower regularization temperatures.

\subsection{Reinforcement Learning with Transformers}\label{sec:architectures}

To represent the policies and value functions for the two self-play processes, Ataraxos employs two transformers~\cite{transformer}, which we call the setup network and the move network. These networks are diagrammed in Extended Data Fig.~\ref{fig:combined-arch}. Important choices included parameterizing the setup network as a decoder-only transformer, which allowed training on entire setups with single forward-backward passes; parameterizing the policy over moves using a key-query matrix product~\cite{monroe}, which learned faster than less sophisticated parameterizations; using learned absolute positional embeddings~\cite{learned-pos} in both networks; and sizing the move network to balance sample efficiency against iteration speed, between which we observed substantial tradeoffs.

\begin{extendeddatafigure}[!htbp]
\centering
\includegraphics[width=\linewidth,height=.82\textheight,keepaspectratio]{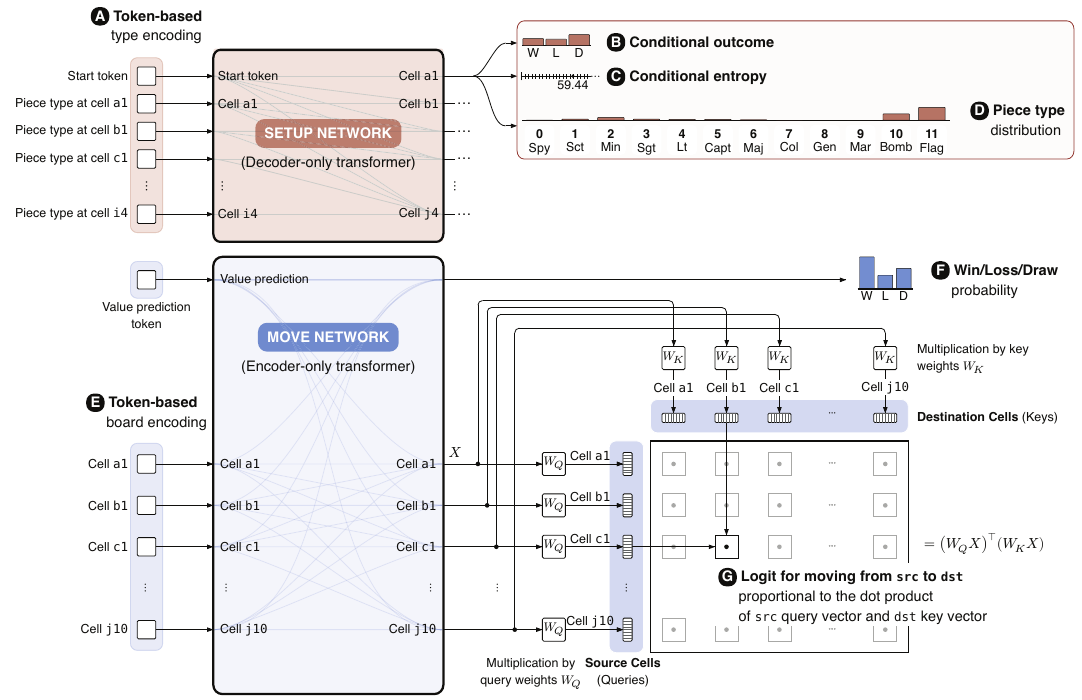}
\setcounter{extendeddatafigure}{0}
\caption{Setup and move network diagrams. \textbf{Top,} the setup network is a decoder-only transformer that generates setups by autoregressively placing the 40 pieces onto squares of the board in row-major order (i.e., first row, first column; first row, second column; and so forth). \textbf{(A)} As input, the setup network takes tokenized piece types with learned absolute positional embeddings~\cite{learned-pos}. As output, for each setup prefix, the setup network provides: \textbf{(B)} an estimate of the conditional probability of winning, losing, or drawing a self-play game; \textbf{(C)} an estimate of the conditional entropy of the setup; and \textbf{(D)} probabilities for placing pieces of each type on that square. \textbf{Bottom,} the move network is an encoder-only transformer that selects the piece to move and the square to which to move it. \textbf{(E)} As input, the move network takes tokenized representations of the occupiable squares with learned absolute positional embeddings, and an additional token for value prediction. The move network outputs: \textbf{(F)} a categorical prediction of the game outcome; and \textbf{(G)} probabilities for the moves computed using a key--query matrix product~\cite{monroe}.}
\label{fig:combined-arch}
\end{extendeddatafigure}

\subsection{Self-Play Training Data Generation}\label{sec:data-generation}

To generate self-play games, Ataraxos selects its setups and moves by directly sampling from the associated networks.

For the moves played during these games, Ataraxos computes the estimates of expected cumulants~\cite{horde} and advantages needed for training updates using $\lambda$-returns~\cite{tdlambda,gae} (with distinct $\lambda$ values), and trains only on the moves with large estimated advantage magnitudes~\cite{robustautonomy}. Filtering in this way reduced the overall wall-clock time per reinforcement learning iteration by a factor of about 2.5, while simultaneously---in a phenomenon meriting further investigation---actually increasing both sample efficiency (with respect to the number of environment queries) and asymptotic performance.

For the setups, Ataraxos uses Monte Carlo returns (i.e., final outcomes of games played by the current policy) to estimate these quantities, and does not employ filtering. The superiority of Monte Carlo returns for advantage estimation is unusual in reinforcement learning, though similar behaviour has also been observed in language model reasoning \cite{r1}.

\subsection{Dynamically Damped Self-Play}\label{sec:damping}

Ataraxos trains on on-policy or close to on-policy self-play data by dynamically damping its learning dynamics. This allows it to leverage standard policy optimization tools to regularize and control the size of its policy updates, obviating the need for more onerous techniques, such as trajectory importance reweighting and policy averaging, for handling imperfect information.

For regularization, Ataraxos incorporates additional terms into the losses of its networks, as detailed under ``Stratego Implementation Details'' in the Supplementary Information. For the setup network, it uses a maximum entropy term \cite{maxentrl}; for the move network, it uses a myopic reverse KL penalty towards the policy that selects a movable piece uniformly at random and then selects a legal move for this piece uniformly at random. Ataraxos anneals the coefficients of these regularization terms according to distinct power laws over training~\cite{mmd}. We found that this regularization functioned somewhat analogously to an energy reserve---annealing too cautiously left the playing ability of the model underdeveloped, while annealing too aggressively produced rapid initial gains but also collapsed the entropy of the model, depleting its capacity to learn thereafter and often making it easy to exploit.

For update size control, Ataraxos employs four mechanisms, which we found offered complementary benefits: a reverse KL penalty to the data collection policy, importance ratio clipping~\cite{ppo}, gradient norm clipping \cite{gradnorm}, and the learning rate of Adam \cite{adam}. Ataraxos anneals the learning rate for its move network according to a power law over the course of training. We found that scheduling this learning rate was crucial both for rapid learning early in training and for preventing plateauing later in training.

\subsection{Belief Modelling}\label{sec:belief-modeling}

To facilitate search, Ataraxos trains a belief network on every position of trajectories sampled from its final self-play policy. Given the information available to the player at a position, this network is trained with teacher forcing to maximize the log-likelihood of the true types of the opponent's hidden pieces. The belief network processes the known information using a transformer encoder (similar to the move network) and predicts the types of unknown pieces autoregressively in row-major order using a transformer decoder. The resulting architecture is diagrammed in Extended Data Fig.~\ref{fig:belief-arch}. Ataraxos applies dropout~\cite{dropout} to the belief network during training to aid generalization to out-of-distribution positions reachable by opponents that play very differently from its setup and move networks, such as humans.

\begin{extendeddatafigure}[!htbp]
\centering
\includegraphics[width=\linewidth,height=.82\textheight,keepaspectratio]{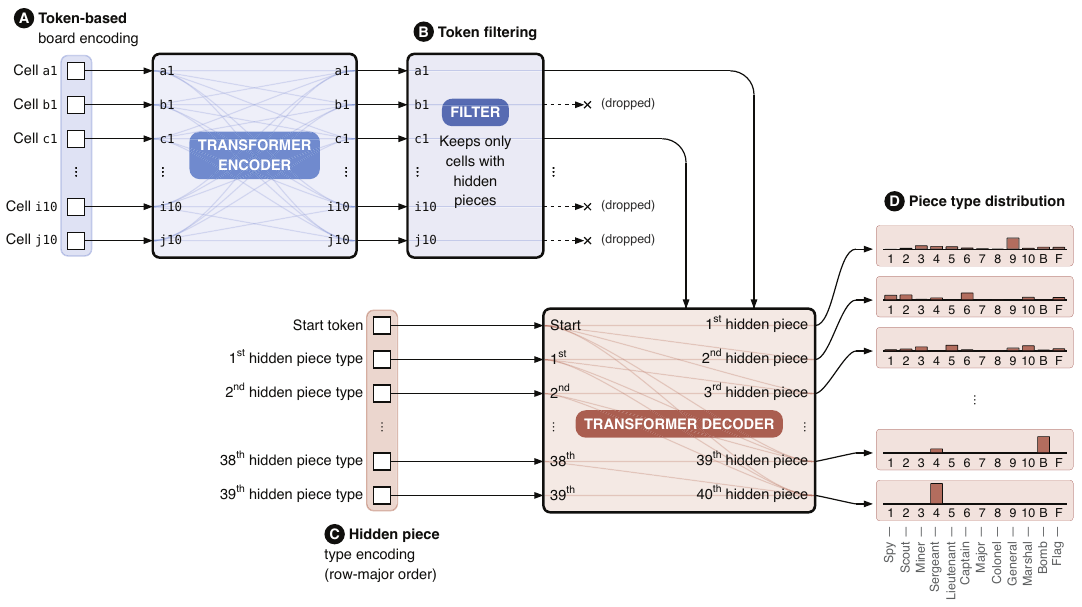}
\setcounter{extendeddatafigure}{3}
\caption{Diagram of belief architecture. \textbf{(A)} The input to the encoder is a tokenized representation of the occupiable squares with learned absolute positional embeddings, similar to the move network. \textbf{(B)} The output of the encoder is filtered to keep only tokens corresponding to squares occupied by hidden pieces of the opponent. Keys and values for these tokens are passed into transformer decoder layers. The inputs \textbf{(C)} and outputs \textbf{(D)} to the decoder are the types and predictions about the types of the opponent hidden pieces, respectively, in row-major order; the inputs use learned positional embeddings.}
\label{fig:belief-arch}
\end{extendeddatafigure}

\subsection{Test-Time Search via Update Equivalence}\label{sec:search}

Given move and belief networks, the search Ataraxos employs is straightforward---it simply performs an additional damped self-play reinforcement learning update~\cite{ues}. That such search is feasible at all, and moreover simple, is noteworthy, as test-time search in settings with as much hidden information as Stratego has been viewed as such a major technical challenge that prior work effectively forwent it~\cite{deepnash}.

Ataraxos uses the belief network prior to each move to sample a collection of possible game states given its current position. For each candidate move, it runs depth-limited rollouts from these game states---starting with that candidate, then using its move network to simulate both players. Ataraxos estimates the value of the candidate move by averaging its network value predictions across the positions reached by rollouts starting from that move. These averaged values approximate self-play action values regardless of the policy of the opponent Ataraxos is facing, since the belief network approximates the posterior distribution of the self-play policy, the rollouts are executed by the self-play policy, and the network predicts self-play position values.

Ataraxos uses these values to update its policy with a tabular step of magnetic mirror descent~\cite{mmd}, which regularizes and controls the size of the update with the same two reverse KL divergences used during training. Importantly, this update can safely be more aggressive than those during training, both because the test-time update is tabular (and thus does not interfere with the policy at other positions), and because it is based on the more accurate advantage estimates enabled by the larger amount of computation per position at test time. The move Ataraxos plays is sampled from the updated policy. Setups, by contrast, are generated by sampling directly from the setup network, with neither search nor any other intervention applied.

\subsection{Training Information}

The reinforcement learning training run utilized 16 H100s for 1 week; the training run for the belief network utilized 4 H100s for 4 days. Extended Data Fig.~\ref{fig:elo} shows metrics for this training run and searches performed on top of its final networks.

\begin{extendeddatafigure}[!htbp]
\centering
\includegraphics[width=\linewidth,height=.82\textheight,keepaspectratio]{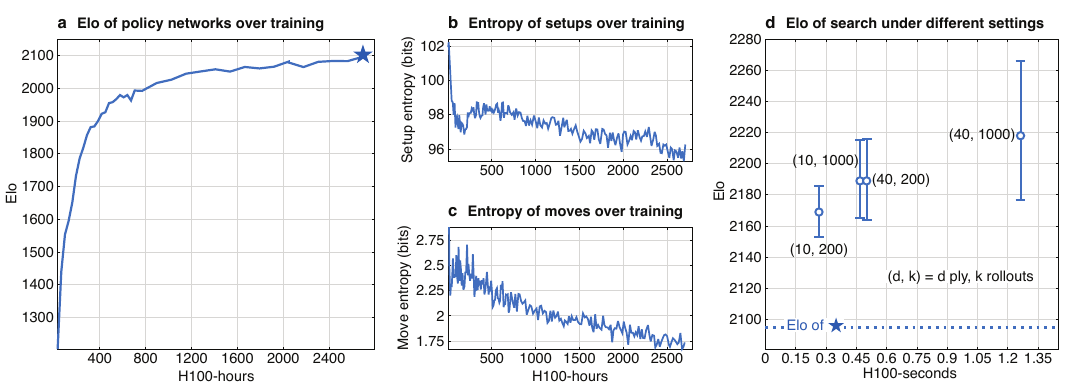}
\setcounter{extendeddatafigure}{1}
\caption{Performance and entropy of Ataraxos as a function of compute. \textbf{a} Elo of the policy networks over the course of training, as measured by performance against a fixed reference opponent. Quantifying skill level in Stratego by Elo is useful but flawed, as randomization can compress margins (i.e., make margins between two players smaller than would be predicted by their respective margins against a common third player whose skill level lies between them), among other reasons. \textbf{b} The entropy of the setups and \textbf{c} the entropy of the moves (in expectation over the self-play distribution of positions). Dynamic damping shepherds these entropies smoothly downward over training, preventing them from collapsing. \textbf{d} Elo of search on a single H100 under varying rollout counts and search depths with 95\% confidence intervals. Deeper searches with more rollouts produce stronger performance but require more time per move. During the evaluation, Ataraxos used a 40-ply 1,000-rollout search that averaged about 1.26 seconds per move---a pace of play faster than that of human players.}
\label{fig:elo}
\end{extendeddatafigure}

The reinforcement learning training run comprised $163$ million finished games, $208$ billion environment steps, $8.56$ million gradient steps for the move network, and $99.5$ thousand gradient steps for the setup network.

Compared to previous work~\cite{deepnash}---which did not reach the level of top humans---our reinforcement learning training run consumed roughly 1/500th of the compute cost, 1/30th of the self-play games, and 1/100th of the training examples. Taken together with the step change in playing strength reported in the main text, the reductions in self-play games and training examples indicate that the lower training cost reflects not only faster implementation but also dramatically greater sample efficiency.

\FloatBarrier
\section{Rules of Stratego} \label{app:gamerules}

\paragraph{Game summary} Stratego is a board wargame played on a 10-by-10 grid with 92 occupiable squares and 2 blocks of non-occupiable squares called lakes  (see Extended Data Fig.~\ref{fig:stratego-rules}a). Each player begins with 40 pieces, arranged in secret on the first four rows of their side so that the identities are concealed from the other player. The game proceeds in alternating turns. On each turn, the acting player moves one piece. If the piece is moved onto a square occupied by a piece of the other player, a battle occurs, at which point both pieces are revealed and at least one of them is removed from the board. Victory is achieved by capturing the Flag of the other player or by leaving the other player with no legal moves; the game ends in a draw if the player to move has no legal moves and the other player would have none were it their turn.

\begin{extendeddatafigure}[!htbp]
\centering
\includegraphics[width=\linewidth,height=.82\textheight,keepaspectratio]{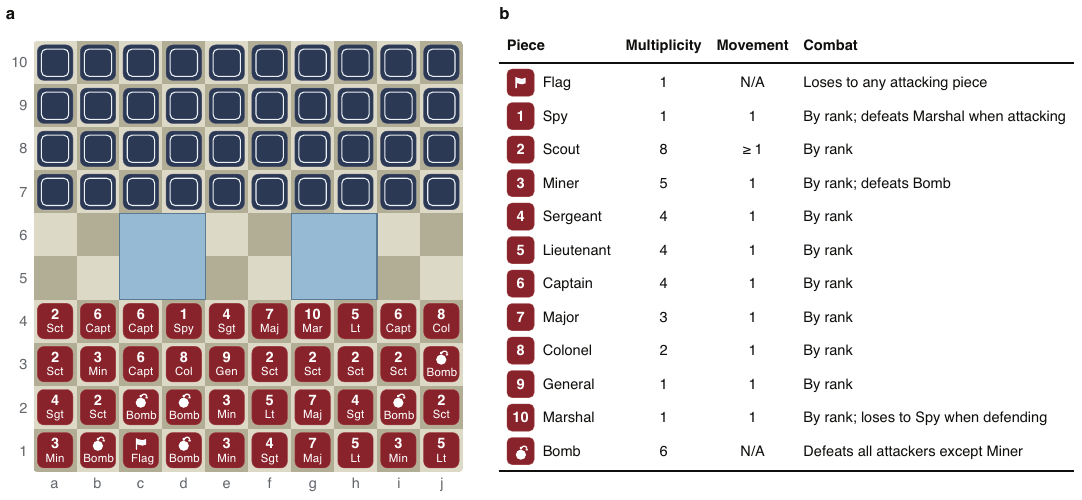}
\setcounter{extendeddatafigure}{2}
\caption{Stratego board layout and piece mechanics. \textbf{a} An example starting position. The light blue blocks in the middle of the board are called lakes and cannot be occupied by the pieces of either player. \textbf{b} Multiplicities, movement abilities, and combat rules for Stratego pieces. Scouts can move an arbitrary number of squares in cardinal directions; other movable pieces can move a single square in cardinal directions. \textit{By rank} means that the higher numbered piece defeats the lower numbered piece.}
\label{fig:stratego-rules}
\end{extendeddatafigure}

\paragraph{Piece details} A description of the pieces is provided in Extended Data Fig.~\ref{fig:stratego-rules}b. Movable pieces---aside from Scouts---can move one square in a cardinal direction (i.e., up, down, left, or right) to squares that are either empty or occupied by an opponent piece; Scouts can move any number of squares in a cardinal direction to squares that are either empty or occupied by an opponent piece, so long as the movement does not jump over a lake or an occupied square. When pieces engage in combat, the outcome is typically determined by rank, which means that the higher-ranking piece defeats the lower-ranking piece if their ranks differ and that both are defeated if their ranks are the same.

\paragraph{Additional rules for competitive play} In competitive play, there are two additional rules. One is the two-square rule, which prohibits a piece from crossing the same square boundary on more than three consecutive turns of its owner. The other is the continuous-chasing rule, which sets limits on the ability of the pieces of one player to chase---in the sense defined below---those of the other.

\paragraph{Threats, evades, chases, chasing}
We define a threat as an action that moves a piece adjacent to a piece of the opponent.
    An evade is an action that moves a piece that was threatened on the previous turn away from the piece that threatened it.
    A chase is an unbroken sequence of alternating threats and evades.
    Finally, chasing is the act of making threats during a chase.

The continuous-chasing rule states that a player who is chasing may not make a threat that would result in a position that has already taken place during the chase, unless that threat would return the moved piece to the square it occupied prior to the previous turn of the chasing player.

\paragraph{Additional rule for online play} There is also an additional rule implemented by Strategus, which is both the primary website for online competition and the website on which the evaluation against Pim Niemeijer was conducted. This rule is called the 200-move rule and states that the game ends in a draw if there is a sequence of 200 moves without a battle.

\paragraph{Time controls} Competitive Stratego also includes time controls. The evaluation against Pim Niemeijer was conducted under the default 15+3 Strategus time controls. The notation 15+3 means that each player starts with a 15-minute buffer and is allocated 3 free seconds for each move before their buffer starts to run down.

\FloatBarrier
\section{Engineering Details}
\label{sec:implementation}

To make the project feasible on modest academic compute infrastructure, we implemented a GPU-accelerated Stratego simulator in CUDA C++. This reduced the runtime of our data collection, training, and search pipelines, while simultaneously reducing memory usage.

Our simulator was designed around five desiderata.
First, it avoids explicitly storing in memory quantities that are easily recomputed. For example, we eliminate the need for a traditional rollout buffer that stores large information states and legal action masks by implementing a simulator capable of travelling back in the history of a game and reconstructing quantities of interest on demand. This greatly reduces memory footprint and fragmentation, as discussed in the Methods section \nameref{app:backend rb}.
Second, it maximizes simulation throughput, aiming for roughly 10 million board state updates per second, while also implementing anti-chasing rules.
Third, it minimizes dynamic memory allocations: the simulator allocates memory as needed at construction time and then manages that memory directly rather than dynamically allocating and deallocating over time.
Fourth, it supports search by enabling fast reset of board states to nonterminal states.
Finally, it treats boards independently: once one game terminates, it is reset independently of whether the other games simulated in parallel have terminated. As a consequence, the boards gradually desynchronize, creating a distribution of training data that covers the different phases of the game.

\subsection{Simulator and Rollout Buffer Design}\label{app:backend rb}

Unlike typical reinforcement learning infrastructure, we do not distinguish between the rollout buffer and the simulator. Rather, we expose a single object, called \verb|StrategoRolloutBuffer|, that serves both purposes. This object tracks a tunable number $N$ of parallel games and is responsible for two critical tasks: supporting historical queries (for example, returning a player's legal action mask or information-state encoding at a past state); and \emph{stepping} the games by receiving actions and updating their states. The latter is performed by calling the \verb|ApplyActions| method, which expects a tensor of $N$ actions as input.

Once a game terminates, a new game starts. A new game is created by sampling a new initial board for both players. The distribution from which the initial board is sampled can be customized. It is also possible to ask the simulator to reset terminated games to a specific \emph{non-initial} game state; this feature is important to support search, as discussed in the Methods section \nameref{app:backend search}.

The rollout buffer tracks game states in a \emph{circular}, preallocated GPU-memory buffer of tunable length. Correspondingly, queries about past states can be supported only if the past state is recent enough.

We found that this design, which integrates aspects of a traditional rollout buffer with a simulator, significantly reduces both memory fragmentation and usage compared to implementing a separate rollout buffer (e.g., using PyTorch). A separate rollout buffer may need to make copies of legal action masks and information state tensors, and pack them into aggregate tensors allocated by that buffer. Instead, our \verb|StrategoRolloutBuffer| is able to reconstruct on-demand past information states and legal action masks. The training loop simply needs to remember at what historical time step the quantity needs to be computed, and then ask the backend to materialize the appropriate tensor to query the value and policy networks. Furthermore, this design is natural when considering that a Stratego simulator \emph{needs} to track history (at least to a nontrivial extent) to implement the anti-chasing rules described in the Methods section \nameref{app:gamerules}. Finally, by endowing the simulator with a notion of history, it becomes possible to implement efficient capturing of past states, as needed to efficiently implement search.

\subsection{Two-Square Rule}\label{app:backend two square}

To efficiently implement the rule, we built a custom state machine whose state is tracked and updated by \verb|StrategoRolloutBuffer|.

The state machine is implemented internally by tracking the last four positions occupied by the last-acting piece for each player. The update logic runs directly on the GPU. Special care needs to be taken to properly account for Scouts, which have special movement abilities.

\subsection{Continuous Chasing Rule}\label{app:backend cc}

To properly implement the rule, the simulator requires access to the board history of each game. To quickly detect whether a move would violate the rule, we used a state machine design paired with a fast diffing algorithm. Custom logic was added to make the rule compatible with resetting terminated games to start from non-initial states (as needed to support search; see the Methods section \nameref{app:backend search}). Indeed, in this case the continuous chasing rule needs to be tested against the history of board states that leads to the non-initial reset state, rather than the history of boards stored in the rollout buffer.

\subsection{Reset Behaviour and Support for Search}\label{app:backend search}

In order to support search, the simulator needs to ``pin'' the simulation of boards to start from a given state. This ability is implemented in our code by asking the \verb|StrategoRolloutBuffer| to reset terminated boards not from an initial state, but rather from a custom non-initial state.

Further implementation details---including the simulator API, information-state representation, training configuration, network architectures, and search hyperparameters---are provided in the Supplementary Information.

\section{Other Experiments}

We show reinforcement-learning ablations in Extended Data Fig.~\ref{fig:training-ablations} and the performance of search across varying hyperparameters in Extended Data Table~\ref{tab:search-abl}. All ablations were performed directly from our final hyperparameter configuration, without retuning the remaining hyperparameters. The results therefore reflect the effect of each ablated component within our final design, rather than the best performance attainable under the ablated design choices.

\begin{extendeddatafigure}[!htbp]
\centering
\includegraphics[width=\linewidth,height=.82\textheight,keepaspectratio]{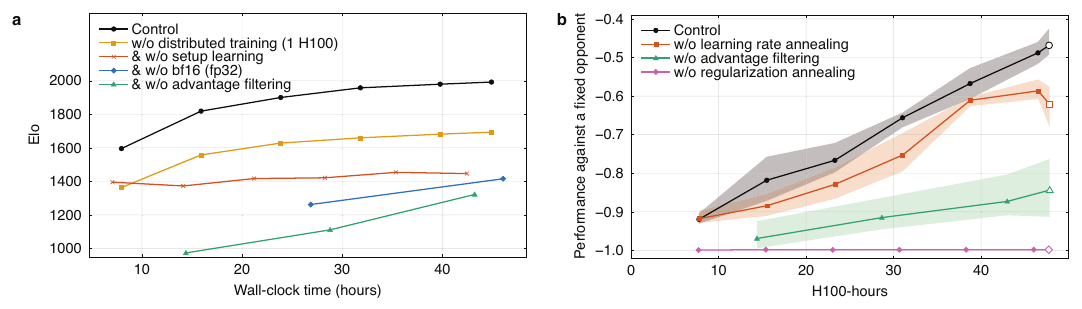}
\setcounter{extendeddatafigure}{4}
\caption{Ablations of the training run. \textbf{a} Single-seed ablations. Without distributed training (yellow), Ataraxos follows a similarly shaped trajectory of improvement over the course of scheduled annealing, but at a substantially lower Elo. Series marked with ``\&'' additionally remove the indicated component. When setup learning is also removed (and replaced by a fixed uniform distribution over setups), the shape of the Elo trajectory flattens---both because of the weakness of uniformly distributed setups and because of the bad assumptions the associated self-play distribution causes the move network to make about the pieces of its opponents. Removing bfloat16 and advantage filtering substantially slows iterations due to more expensive network calls and larger training datasets, respectively. Removing advantage filtering also causes a large increase in move entropy and large decrease in sample efficiency. \textbf{b} Ablations of learning-rate annealing, advantage filtering, and regularization annealing across three seeds. Lines show the mean across seeds and shading shows the minimum and maximum seed performance. All three ablated components materially affect performance over the course of training.}
\label{fig:training-ablations}
\end{extendeddatafigure}

\begin{extendeddatatable}[!htbp]
\centering
\includegraphics[width=\linewidth,height=.82\textheight,keepaspectratio]{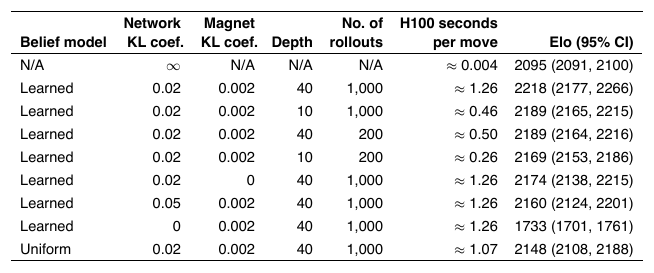}
\setcounter{extendeddatatable}{0}
\caption{Elo of search with varying hyperparameters. The first row shows direct sampling from the move network (i.e., no search); the second shows the search that was used in the evaluation. Both Elo and time expended increase with the depth of the search and the number of rollouts. Reverse KL regularization strength has a large effect on performance. Removing the reverse KL term to the move network devastates performance---causing it to drop well below that of the move network---because the search overfits to idiosyncrasies of the move network that fail to generalize to other opponents. The final row ablates the learned belief model, replacing it with the belief associated with the uniformly random policy.}
\label{tab:search-abl}
\end{extendeddatatable}

We additionally compared the performance of the current iterates and their exponential moving averages over training across three seeds by evaluating each against a fixed reference model. As shown in Extended Data Fig.~\ref{fig:EMA}, the exponential moving average achieved similar or better mean performance while exhibiting lower variation across seeds.

\begin{extendeddatafigure}[!htbp]
\centering
\includegraphics[width=\linewidth,height=.82\textheight,keepaspectratio]{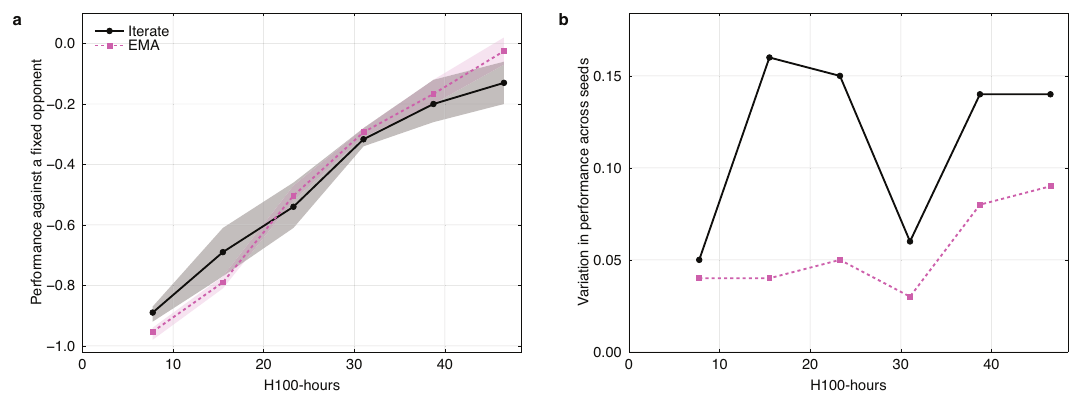}
\setcounter{extendeddatafigure}{5}
\caption{Comparison of current iterates and exponential moving averages across three seeds. \textbf{a} Performance against a fixed reference model over training, with the line showing the mean across seeds and the shading showing the minimum and maximum seed performance. \textbf{b} Difference at each iteration between the strongest and weakest seed. The exponential moving average attains similar or better mean performance while exhibiting lower variation across seeds.}
\label{fig:EMA}
\end{extendeddatafigure}

We also evaluated Ataraxos against the bots from the Stratego Evaluator benchmark~\cite{strategoevaluator}: Asmodeus, Celsius, Celsius1.1, Vixen, and Peternlewis. Ataraxos won 99, 98, 97, 99, and 98 of 100 games against these opponents, respectively. Win percentages this close to 100 imply that the Stratego Evaluator benchmark is meaningful only as a sanity check.

\FloatBarrier
\section{Learned Setups}

The probability distribution for each piece in a setup chosen by Ataraxos is given in Extended Data Fig.~\ref{fig:setup-probabilities}, assuming that the Flag is on the left side of the setup. The probabilities for the right side are symmetric, as Ataraxos applies a left-right orientation randomization to its setups after sampling from the setup network to enforce symmetry. Ataraxos uses a bombed-in Flag (i.e., a Flag that is enclosed by Bombs, such as in Extended Data Fig.~\ref{fig:stratego-rules}a) about two-thirds of the time. The setups Ataraxos used in the series against Pim Niemeijer were generated fully autonomously by sampling from this distribution, one per game, and are shown in Extended Data Fig.~\ref{fig:pim-setups}.

\begin{extendeddatafigure}[!htbp]
\centering
\includegraphics[width=\linewidth,height=.82\textheight,keepaspectratio]{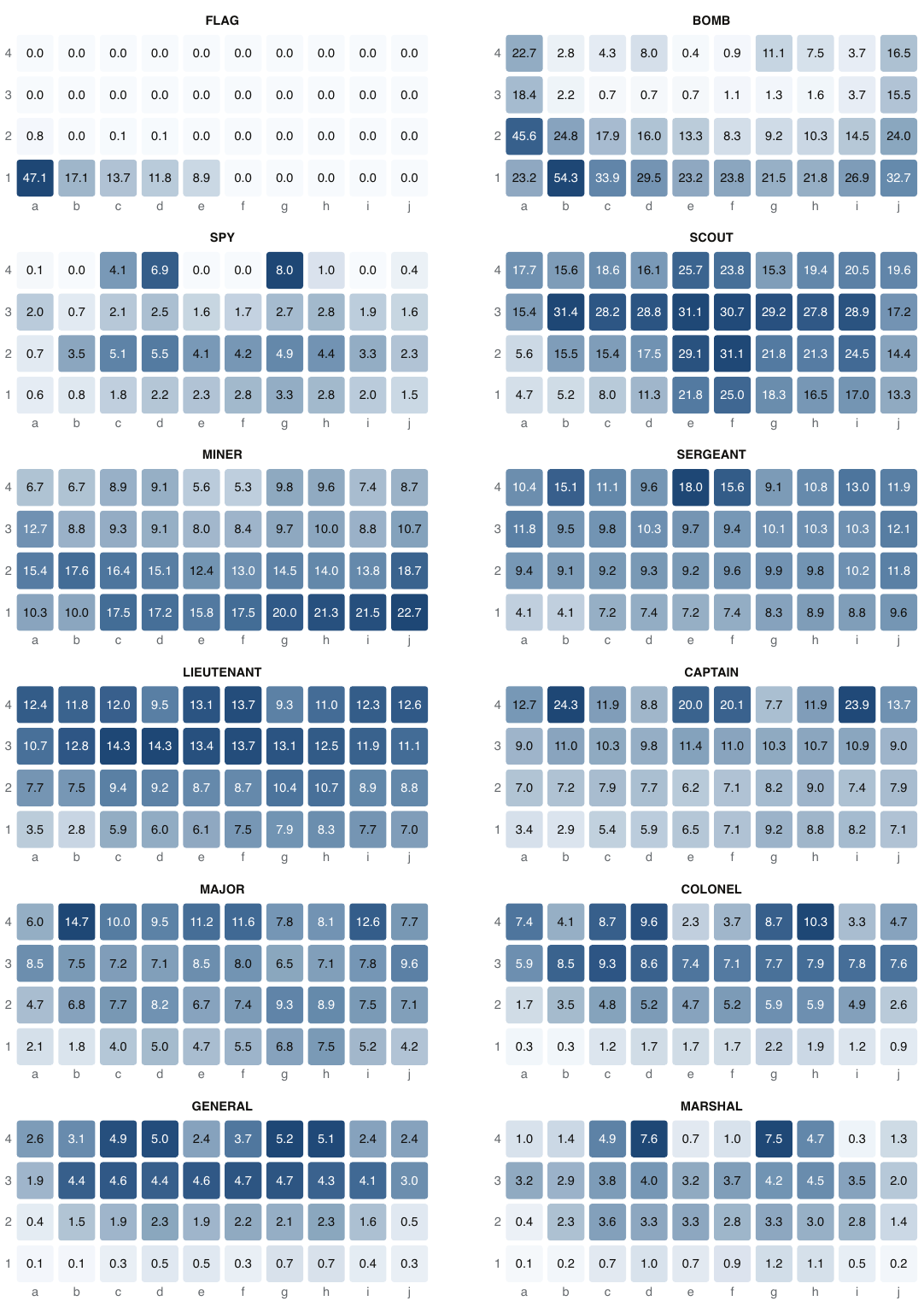}
\setcounter{extendeddatafigure}{6}
\caption{Estimated probabilities (in percent) of pieces being different types in setups chosen by Ataraxos, assuming that the Flag is on the left side of the setup.}
\label{fig:setup-probabilities}
\end{extendeddatafigure}

\begin{extendeddatafigure}[!htbp]
\centering
\includegraphics[width=\linewidth,height=.82\textheight,keepaspectratio]{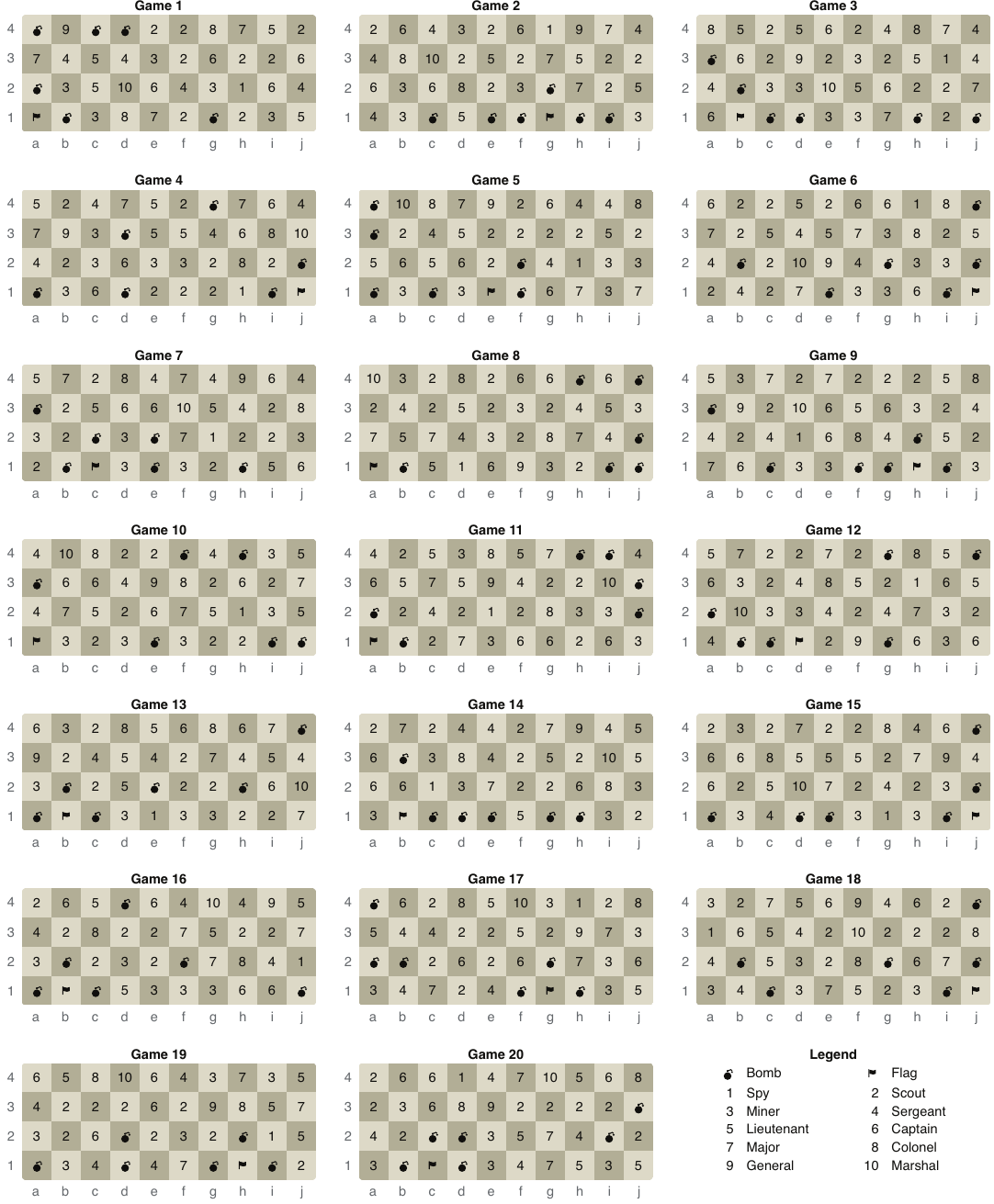}
\setcounter{extendeddatafigure}{7}
\caption{The setups Ataraxos played in the series against Pim Niemeijer.}
\label{fig:pim-setups}
\end{extendeddatafigure}

\FloatBarrier
\section{Play Style}

The play style of Ataraxos differs from those of top human players (which are discussed, for example, by de~Boer~\cite{deBoer2007Invincible}), in terms of both setups and gameplay.

\subsection{Setups}

Players observed that Ataraxos uses aggressive setups (i.e., ones in which high-value pieces are at or near the front), high Bombs (i.e., Bombs in the 3rd and 4th rows), and back corner Flags (which humans consider difficult to defend) more frequently than humans.
They also observed that it generally uses setups that have less predictable structure than those of humans.

\subsection{Gameplay}

Players observed the following. Ataraxos moves in a manner that makes its pieces hard to read (i.e., discern the types of) relative to the movement patterns of human players.
   Ataraxos is more willing than humans to accept a draw during the opening when progressing the game would have negative expected value (as happened in Game 11 against Pim Niemeijer).
   Ataraxos has a stronger preference than humans for preserving Scouts deep into the game.
   Ataraxos uses certain bluffs sparingly relative to humans, but uses other bluffs that humans consider altogether too risky to play.
   Ataraxos fights more bitterly than humans when it is behind, aggressively stalling and pestering to slow the progression of its opponent---tactics some humans consider rude.
   Ataraxos excels relative to humans at long-term positional play, punishing mistakes, defending, playing from an information deficit, transitioning from middle- to end-game, and negotiating favourable positions into wins and unfavourable positions into wins or draws.
   Ataraxos takes gambles that humans consider arrogant (in the sense of not respecting the opponent).
   Ataraxos feels preternaturally lucky, always seeming to have the pieces it needs in the right places, to have its gambles pay off, and to have its opponents do \textit{as it wants}.

\section{Contextualization vis-à-vis DeepNash}

DeepNash~\cite{deepnash} is an AI for Stratego that was developed by DeepMind.

\subsection{Evaluation} DeepNash was evaluated on the website Gravon in April 2022, winning 42 of the 50 games counted by Perolat et al.~\cite{deepnash}, but not achieving the top ranking on the site. Several aspects of this evaluation are relevant to interpreting its results:
1) By the time of the evaluation, the player base had largely moved away from Gravon (only 25 players are listed for the final ranking of 2022~\cite{Gravon2022StrategoRating})---as a result, the opponents matched against DeepNash were far from the level of top humans~\cite{gravon_starship}; 2) the one-off online match setting may not have elicited the maximum effort level from the human opponents, who were not aware that an official evaluation was taking place; 3) these opponents had no reason to look for anti-bot exploits, as they were not aware that they were playing against a bot; and 4) these opponents were not aware that DeepNash played a fixed strategy (and thus did not know that it was safe to exploit the same weakness across multiple games).

Separately, at the 2023 Stratego World Championship, DeepNash was demoed against human players, recording 19 wins and 9 losses~\cite{deepnash_wc2023}; DeepNash lost to most of the highest-ranked players who played against it, including Pim.

We reached out to DeepMind to ask if they would allow an evaluation between DeepNash and Ataraxos, offering to build any infrastructure necessary for the evaluation. DeepMind responded that it would not be possible as the code for DeepNash is no longer functional.

\subsection{Compute Cost}\label{app:compute-cost} Perolat et al.~\cite{deepnash} state that DeepNash was trained on $1{,}024$ TPU nodes. To the recollection of the corresponding author of DeepNash with whom we spoke, this training took between 2 and 3 months and used TPU v3s. Under 2025 pricing~\cite{googlecloud_tpu_pricing_2025}, such a training run would cost roughly between \$3,000,000 and \$4,500,000, depending on how much of the third month was used.

The reinforcement learning models and belief models of Ataraxos were trained on 16 H100s for 1 week and 4 H100s for 4 days, respectively. Such a run costs less than \$8,000 at 2025 prices~\cite{VoltageParkPricing}.

\subsection{Sample Cost}
The training run for DeepNash consumed about 5.5 billion games and between 5 and 10 trillion training examples.

The training run for Ataraxos consumed about 160 million games and about 50 billion training examples.

\section{Opportunities for Further Improvement}

\paragraph{Learning} Ataraxos accesses history through features rather than learning across time directly. For the belief model, we found that much stronger compute-normalized performance could be attained by interleaving spatial attention with either temporal attention \cite{spatiotemporal} or recurrent models. However, for reinforcement learning, we did not observe an analogous out-of-the-box compute-normalized performance improvement due to the additional runtime and memory requirements of such architectures, as well as their interplay with advantage filtering. We believe such architectures could achieve stronger performance given a sufficiently large compute budget or provided with additional runtime, memory, or design optimizations.

\paragraph{Search} The amount of improvement attainable by the search procedure of Ataraxos is ultimately bounded because it is mimicking a single update step. A more sophisticated search algorithm would be able to leverage arbitrary amounts of additional compute to continue to improve the policy. One possible route towards this end would be to incorporate innovations from knowledge-limited subgame solving~\cite{obscuro25}.

\section{Barrage Stratego Evaluation Details}

The evaluations took place on Strategus using the default 5+1 time controls (meaning that each player starts with a 5-minute buffer and is allocated 1 free second for each move before their buffer starts to run down).
Games were scheduled by the players at their convenience.
Prior to the start of the evaluation, the players were informed that Ataraxos would not adapt to their play and that they would be evaluated by effective win rate.

Because belief model training was ongoing at the time at which the evaluation began, we first evaluated the policy network (without search). The policy network won three series:  29 wins, 17 losses, and 4 draws (a large margin) against world \#4 Axel Hangg; 36 wins, 12 losses, and 2 draws (a very large margin) against world \#3 Sébastien Crot; and 26 wins, 21 losses, and 3 draws against world \#1 Pim Niemeijer. We evaluated the search policy in a second 50-game series against Pim, chosen because he had lost to the policy network by the smallest margin. This second series put the search policy at a disadvantage: by the time Pim played it, he had already accumulated information about the setup policy, which the two policies shared, and about their otherwise similar playstyles. Nonetheless, the search policy substantially outperformed the policy network, winning the series with 31 wins, 14 losses, and 5 draws.

Implementation and training details are reported in the ``Barrage Stratego Implementation Details'' section of the Supplementary Information.

\section{Hanabi Evaluation Details} \label{sec:hanabi-perf}

We evaluated Ataraxos on the two- to five-player variants of Hanabi, with $\{0, 1, \ldots, N\}$ players running search in the $N$-player variant to demonstrate its power as a scalable \emph{multi-agent} search method. Training, network, and search details are reported in the ``Hanabi Implementation Details'' section of the Supplementary Information.

The evaluation results are summarized in Extended Data Fig.~\ref{fig:hanabi-results}, with detailed numbers listed in Supplementary Table~\hyperref[tab:hanabi-numbers]{S32}. All values are means $\pm$ standard errors computed over $10{,}000$ games for each condition. With search employed by all players, Ataraxos achieves the following average scores and percentages of perfect (25-point) games:
\begin{itemize}
    \item 2 players: score $24.654 \pm 0.007$, perfect games $77.53\% \pm 0.42\%$
    \item 3 players: score $24.863 \pm 0.005$, perfect games $89.90\% \pm 0.30\%$
    \item 4 players: score $24.852 \pm 0.005$, perfect games $88.40\% \pm 0.32\%$
    \item 5 players: score $24.410 \pm 0.009$, perfect games $58.05\% \pm 0.49\%$
\end{itemize}

\begin{extendeddatafigure}[!htbp]
\centering
\includegraphics[width=\linewidth,height=.82\textheight,keepaspectratio]{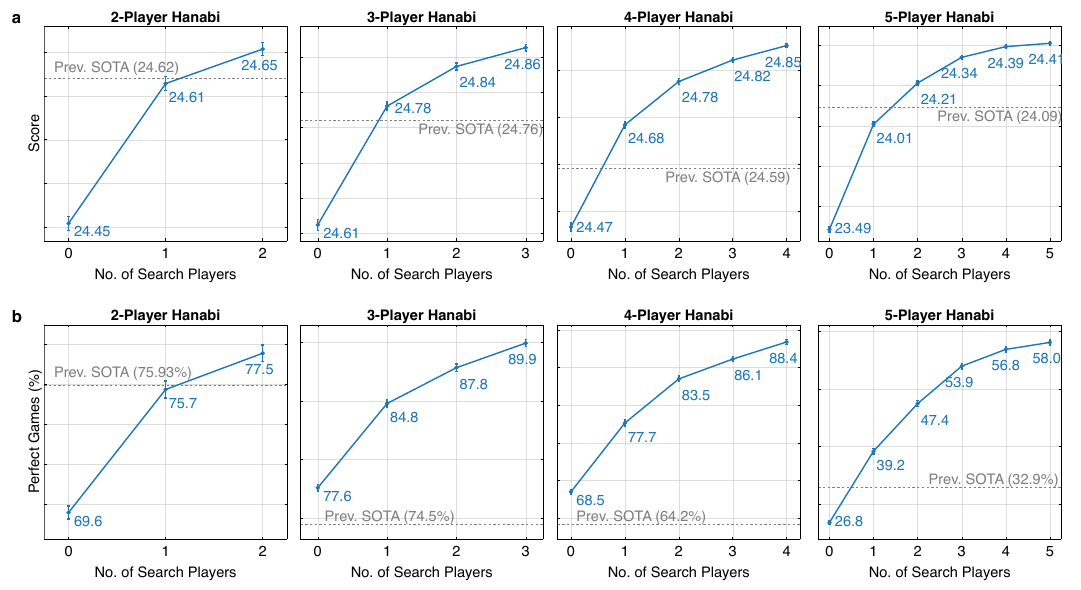}
\setcounter{extendeddatafigure}{8}
\caption{Performance of Ataraxos on Hanabi over $10{,}000$ games with standard errors. \textbf{a} Average score. \textbf{b} Percentage of perfect (25-point) games.}
\label{fig:hanabi-results}
\end{extendeddatafigure}
In Extended Data Fig.~\ref{fig:hanabi-results}a, we plot the performance in terms of average scores over $10{,}000$ games and compare these results to previous state-of-the-art results from Fickinger et al.~\cite{fickinger2021scalable} for the two-player variant and Forkel et al.~\cite{forkel2026highentropyleadssymmetry} for the three-, four-, and five-player variants.
Importantly, performance improves monotonically as the number of players using search increases, with particularly large gains for variants with more players as a result of applying Ataraxos search to multiple agents.
The gains in the average scores may seem numerically small; however, because Hanabi scores are capped at 25 points and additional points become increasingly difficult to secure near this ceiling, these results reflect substantial progress. Another way to understand this progress is through the percentage of perfect games that reach 25 points. In Extended Data Fig.~\ref{fig:hanabi-results}b, we observe large increases in the percentage of perfect games across variants as we apply Ataraxos search to more players.

\FloatBarrier
\section{\textit{Dou Dizhu} Evaluation Details}\label{sec:doudizhu-perf}

We further evaluated Ataraxos on \textit{dou dizhu}. The reinforcement learning, belief-model, and search details are reported in the ``\textit{Dou Dizhu} Implementation Details'' section of the Supplementary Information.

We compared Ataraxos against the previous state-of-the-art method, PerfectDou~\cite{yang2022perfectdou}, as well as its predecessor, DouZero~\cite{zha2021douzero}. We evaluated heads-up performance in a duplicated setting: one agent plays as the landlord, while the opposing agent controls the two independent peasants. Performance is reported as role-averaged score, computed by comparing the same pair of agents after swapping their roles. We evaluated Ataraxos against each baseline over $10{,}000$ duplicated deals and report the performance in Supplementary Table~\hyperref[tab:doudizhu-performance]{S33}. All values are means $\pm$ standard errors; comparisons involving Ataraxos use $10{,}000$ duplicated deals, whereas comparisons among the policy network, PerfectDou, and DouZero use $100{,}000$ duplicated deals. The role-averaged scores of each agent against the others are as follows:
\begin{itemize}
    \item Ataraxos: $0.107 \pm 0.012$ versus the policy network, $0.199 \pm 0.015$ versus PerfectDou, $0.350 \pm 0.016$ versus DouZero
    \item Policy network (i.e., Ataraxos without search): $-0.107 \pm 0.012$ versus Ataraxos, $0.152 \pm 0.004$ versus PerfectDou, $0.286 \pm 0.004$ versus DouZero
    \item PerfectDou: $-0.199 \pm 0.015$ versus Ataraxos, $-0.152 \pm 0.004$ versus the policy network, $0.142 \pm 0.004$ versus DouZero
    \item DouZero: $-0.350 \pm 0.016$ versus Ataraxos, $-0.286 \pm 0.004$ versus the policy network, $-0.142 \pm 0.004$ versus PerfectDou
\end{itemize}
Ataraxos outperformed existing methods by a large margin and achieved the strongest performance against each baseline, establishing a new state of the art.

As a supplementary analysis, we evaluated Ataraxos under different update step sizes using the role-averaged score against PerfectDou. We report the results in Supplementary Figure~\hyperref[fig:doudizhu-update-sizes]{S8}. To reduce evaluation cost, this sweep was run with a lighter schedule of $2{,}500$ duplicated deals per setting.

\begingroup
\raggedright
\renewcommand{\refname}{Methods References}

\endgroup

\section*{Data availability}

No static training datasets were used in this project; all training data were generated from scratch by the agents as they learned. Static records of the 20 evaluation games against Pim Niemeijer are available at \url{https://ataraxosai.github.io}.

\section*{Code availability}

Code used for the Stratego, Hanabi, and \textit{dou dizhu} experiments is publicly available at \url{https://github.com/AtaraxosAI}.

\section*{Acknowledgements}

We thank the site administrator of Strategus for allowing us to evaluate Ataraxos via Strategus. We thank Winner, Rayan Manji, and Rein Halbersma for playtesting, feedback, and other assistance. We thank Hani Basilious, Shenglong Wang, and NYU's high performance computing team for providing and facilitating the compute for the project. We thank Sara Werner for helping to facilitate the evaluation. We thank Joseph Twin for bringing expositional errors to our attention.

\section*{Funding}

S.S. discloses support for the research of this work from the Office of Naval Research awards N000142212121 and N000142512116.
E.V. discloses support for the research of this work from the NYU Department of Civil and Urban Engineering startup funding, and 2SMART Center, U.S. Department of Transportation Grant No. 69A3552348326, and U.S. National Science Foundation award IIS-2552047.
H.H. declares no relevant funding.
Z.F. discloses support for the research of this work from the National Science Foundation award CCF-2443068, the Office of Naval Research award N000142512296, and a Schmidt Sciences AI2050 Early Career Fellowship.
J.Z.K. discloses support for the research of this work from the Office of Naval Research awards N000142212121 and N000142512116.
G.F. discloses support for the research of this work from the National Science Foundation awards CCF-2443068 and IIS-2552046, the Office of Naval Research award N000142512296, and a Schmidt Sciences AI2050 Early Career Fellowship.

\section*{Author contributions}

S.S.: Conceptualization, Data curation, Formal analysis, Investigation, Methodology, Project administration, Software, Supervision, Validation, Visualization, Writing --- original draft, Writing --- review \& editing.
E.V.: Data curation, Formal analysis, Funding acquisition, Investigation, Methodology, Resources, Software, Validation, Visualization, Writing --- review \& editing.
H.H.: Data curation, Formal analysis, Investigation, Methodology, Software, Validation, Visualization, Writing --- review \& editing.
Z.F.: Data curation, Formal analysis, Investigation, Methodology, Software, Validation, Visualization, Writing --- review \& editing.
J.Z.K.: Conceptualization, Funding acquisition, Methodology, Resources, Writing --- review \& editing.
G.F.: Conceptualization, Data curation, Formal analysis, Funding acquisition, Investigation, Methodology, Project administration, Resources, Software, Supervision, Validation, Visualization, Writing --- original draft, Writing --- review \& editing.

\section*{Competing interests}

The authors declare no competing interests.

\section*{Ethics statement}

The Carnegie Mellon University Institutional Review Board (CMU IRB) determined that the human-player evaluation study qualified for exemption under the 2018 Common Rule, 45 CFR 46.104(d)(3)(i)(B), as a low-risk benign behavioural intervention (STUDY2025\_00000013; modification MOD202500000509). Informed consent was obtained from all participants in accordance with the CMU IRB-reviewed consent procedures.

\section*{Correspondence}

Correspondence and requests for materials should be addressed to Samuel Sokota or Gabriele Farina. Email: \href{mailto:ssokota@andrew.cmu.edu}{ssokota@andrew.cmu.edu} (S.S.) or \href{mailto:gfarina@mit.edu}{gfarina@mit.edu} (G.F.).

\section*{Additional information}

\textbf{Supplementary information} is available for this paper.

\clearpage
\part*{Supplementary Information}
\setcounter{secnumdepth}{3}
\setcounter{section}{0}
\setcounter{table}{0}
\setcounter{figure}{0}
\renewcommand{\thesection}{S\arabic{section}}
\renewcommand{\thetable}{S\arabic{table}}
\renewcommand{\thefigure}{S\arabic{figure}}
\renewcommand{\theHsection}{S\arabic{section}}
\renewcommand{\theHtable}{S\arabic{table}}
\renewcommand{\theHfigure}{S\arabic{figure}}

The sections below provide detailed implementation, training, and evaluation material supporting the Methods and Extended Data figures and tables.

\section{Simulator API}\label{app:si-api}
The following table describes the main historical queries supported by the simulator, complementing the Methods subsection \enquote{Simulator and Rollout Buffer Design}.
{\small
\rowcolors{2}{gray!10}{white}
\noindent\begin{longtable}{@{}L{0.39\linewidth}L{0.10\linewidth}L{0.14\linewidth}L{0.29\linewidth}@{}}
\caption{Historical simulator queries exposed by the API.}\label{tab:simulator-api}\\
\em Method & \em Input & \em Output & \em Description\\
\toprule
\endfirsthead
\em Method & \em Input & \em Output & \em Description\\
\toprule
\endhead
    \texttt{CurrentStep} & (none) & 64-bit integer & Returns the number of state updates performed by the simulator.\\
    \texttt{ActingPlayer} & time step & 0 or 1 & The player acting at the current time step (0 for the red player, 1 for the blue player).\\
    \texttt{ComputeLegalActionMask} & time step & (none) & Computes the legal action mask for the acting player at the given time step. The answer is written in a preallocated tensor kept by the rollout buffer. \\
    \texttt{ComputeInfostateTensor} & time step & (none) & Computes the information state for the acting player at the given time step. The answer is written in a preallocated tensor kept by the rollout buffer. \\
    \texttt{ComputeRewardPl0} & time step & (none) & Computes the reward for the red player at the given time step. The reward is set to 0 for those boards that are not terminated at the given time step. The answer is written in a preallocated tensor kept by the rollout buffer.\\
    \texttt{ComputeIsUnknownPiece} & time step & (none) & Computes a boolean tensor indicating which pieces are unknown to the acting player. The answer is written in a preallocated tensor kept by the rollout buffer.\\
    \texttt{ComputePieceTypeOnehot} & time step & (none) & Computes a one-hot encoding of the piece types on the board from the perspective of the acting player. The answer is written in a preallocated tensor kept by the rollout buffer.\\
    \texttt{ComputeTwoSquareRuleApplies} & time step & (none) & Computes whether the two-square repetition rule applies at the given time step. The answer is written in a preallocated tensor kept by the rollout buffer.\\
    \texttt{GetTerminatedSince} & time step & Tensor & For each game, computes the number of steps since it terminated. For non-terminated games, this value is 0. Returns a non-owning tensor.\\
    \texttt{GetHasLegalMovement} & time step & Tensor & Computes whether the acting player has at least one legal move. Returns a non-owning tensor.\\
    \texttt{GetFlagCaptured} & time step & Tensor & Computes whether a player has captured the opponent's flag.  Returns a non-owning tensor.\\
    \texttt{GetPlayedActions} & time step & Tensor & Retrieves the action taken at the specified time step for each game. Returns a non-owning tensor.\\
    \texttt{GetNumMoves} & time step & Tensor & Gets the total number of moves played in each game. Returns a non-owning tensor.\\
    \texttt{GetNumMovesSinceLastAttack} & time step & Tensor & Gets the number of moves since the last attack for each game.  Returns a non-owning tensor.\\
    \texttt{BoardStrs} & time step & Vector of strings & Returns a string representation of the board state for each game at the current time step. Useful for debugging.\\
    \texttt{SampleRandomLegalAction} & GPU-allocated tensor & (none) & Fills out the provided GPU-allocated tensor with a legal action drawn uniformly for each game at the current step.\\
    \bottomrule
\end{longtable}}

\section{Stratego Information State Representation} \label{app:infostate}

The representation of the information state for each player comprises two components:
\begin{itemize}
    \item A representation of the current \emph{board state}; and
    \item An encoding of the past $K$ moves.
\end{itemize}
We describe each component in detail in the following subsections.

\subsection{Representation of the Board State}
\label{app:board-state}

We encode information about the state of the board and the history of the game as a collection of channels, each of which is a $10\times 10$ matrix of floating point numbers. The channels can be divided into different logical subgroups, as follows.

\paragraph{Basic piece properties} The following channels expose basic information regarding the pieces on the board, including their visibility and whether they have moved in the past.
\begin{channeltable}
0--11 & Indicator of whether the piece occupying each square belongs to the player and has type $T$, for each of the 12 types $T \in \{$Spy, Scout, Miner, Sergeant, Lieutenant, Captain, Major, Colonel, General, Marshal, Flag, Bomb$\}$.\\
12--23 & For each square, probability that it belongs to the opponent and has type $T$ for each of the 12 types $T \in \{\text{Spy},\dots,\text{Bomb}\}$, assuming the setup and moves were selected uniformly at random.\\
24--35 & Same as channels 12--23, but from the point of view of the opponent player.\\
\midrule
36 & Indicator of what squares on the board are occupied by hidden pieces of the player.\\
37 & Indicator of what squares on the board are occupied by hidden pieces of the opponent.\\
38 & Indicator of what squares on the board are empty.\\
39 & Indicator of what squares on the board are occupied by pieces of the player that moved.\\
40 & Indicator of what squares on the board are occupied by pieces of the opponent that moved.\\
41 & Fraction of the maximum number of moves per game (see Section~\ref{app:game rules}) that have been exhausted.\\
42 & Fraction of the maximum number of moves between attacks (see Section~\ref{app:game rules}) that have been exhausted.\\
\end{channeltable}

\paragraph{Threats, evasions, attack opportunities} The following channels encode information regarding the historical behaviour of pieces, including whether pieces threatened or evaded other pieces, and whether pieces opted not to attack other pieces when they had the opportunity.

For the purposes of the next channels, we recall that a \emph{threat} is a movement of one's piece so as to make the piece adjacent to a piece of the opponent. The opponent's piece is considered \emph{threatened}.
\begin{channeltable}
43--52 & Indicator of whether the piece occupying the square belongs to the player, and has ever threatened a piece of the opponent which, at the time of the threat, was a revealed piece of type $T \in \{\text{Spy},\dots,\text{Marshal}\}$.\\
53 & Indicator of whether the piece occupying the square belongs to the player, and has ever threatened a piece of the opponent which, at the time of the threat, was hidden.\\
\end{channeltable}

An \emph{evasion} is realized when a piece that was threatened in the immediately preceding move is moved away from the threatening piece, to an empty square.
\begin{channeltable}
54--63 & Indicator of whether the piece occupying the square belongs to the player, and has ever evaded a threat from an opponent's piece which, at the time of the threat, was a revealed piece of type $T \in \{\text{Spy},\dots,\text{Marshal}\}$.\\
64 & Indicator of whether the piece occupying the square belongs to the player, and has ever evaded a threat from an opponent's piece which, at the time of the threat, was hidden.\\
\end{channeltable}

Finally, we define a piece to be \emph{actively adjacent} to an opponent's piece when the player controlling the former has had the chance to attack the latter, but opted not to.
\begin{channeltable}
65--74 & Indicator of whether the piece occupying the square belongs to the player, and has ever been actively adjacent to an opponent's piece which, at the time of the adjacency, was a revealed piece of type $T \in \{\text{Spy},\dots,\text{Marshal}\}$.\\
75 & Indicator of whether the piece occupying the square belongs to the player, and has ever been actively adjacent to an opponent's piece which, at the time of the adjacency, was hidden.\\
\end{channeltable}

The same information regarding threats, evasions, and active adjacencies is then reported from the point of view of the opponent.
\begin{channeltable}
76--85 & Same as channels 43--52, but from the point of view of the opponent player.\\
86 & Same as channel 53, but from the point of view of the opponent player.\\
87--96 & Same as channels 54--63, but from the point of view of the opponent player.\\
97 & Same as channel 64, but from the point of view of the opponent player.\\
98--107 & Same as channels 65--74, but from the point of view of the opponent player.\\
108 & Same as channel 75, but from the point of view of the opponent player.\\
\end{channeltable}

\paragraph{Captured pieces} The next group of channels encodes the starting positions of the pieces that have been captured.
\begin{channeltable}
109--119 & Indicator of whether the given square was occupied, at the start of the game, by a piece of the player of type $T \in \{\text{Spy},\dots, \text{Marshal}, \text{Bomb}\}$ that has now been captured.\\
120--130 & Indicator of whether the given square was occupied, at the start of the game, by a piece of the opponent of type $T \in \{\text{Spy},\dots, \text{Marshal}, \text{Bomb}\}$ that has now been captured.\\
\end{channeltable}

\paragraph{Causes of death} The following channels encode the locations in which pieces died, as well as what interaction caused their death. The following causes of death are distinguished:
\begin{enumerate}
    \item The piece attacked an opponent piece of strictly stronger rank, which was revealed at the time of the attack.
    \item The piece attacked an opponent piece of equal rank, which was revealed at the time of the attack.
    \item The piece attacked an unrevealed opponent piece, and lost the battle.
    \item The piece was revealed, and was attacked by an opponent piece (revealed or not). It lost the attack because it was strictly weaker.
    \item The piece was revealed, and was attacked by an opponent piece (revealed or not). The battle was a tie.
    \item The piece was unrevealed, and was attacked by an opponent piece (revealed or not).
\end{enumerate}

\begin{channeltable}
131--140 & Whether a piece of the player of type $T\in\{\text{Spy},\dots, \text{Marshal}\}$ died in this square, due to cause of death 1.\\
141--150 & Whether a piece of the player of type $T\in\{\text{Spy},\dots, \text{Marshal}\}$ died in this square, due to cause of death 2.\\
151--180 & Whether a piece of the player of type $T\in\{\text{Spy},\dots, \text{Marshal}, \text{Bomb}\}$ died in this square, due to causes of death 3--5 (channels 151--160, 161--170, and 171--180, respectively).\\
181--190 & Whether a piece of the player of type $T\in\{\text{Spy},\dots, \text{Marshal}, \text{Bomb}\}$ died in this square, due to cause of death 6.\\
191--250 & Same as channels 131--190, but for the opponent.\\
\end{channeltable}

\paragraph{Protection moves}
The final group of channels refers to protection moves. When a player's piece $A$ threatens a piece $B$ of the opponent, the opponent might respond by moving one of their pieces, say $C$, to become adjacent to $B$. This way, if the player follows through with the aggression, the opponent can attack $A$ with $C$. Moving $C$ to become adjacent to $B$ as a response to a threat from $A$ is a \emph{protection} move. More precisely, we say that piece $C$ protected piece $B$ against piece $A$; equivalently, piece $B$ was protected by piece $C$ against piece $A$. We extend the same terminology to the case in which $B$ is an empty square.

\begin{channeltable}
251--261 & Whether the piece in the square belongs to the player, and has ever protected a piece of the player which, at the time of the protection, had revealed type $T \in \{\text{Spy},\dots,\text{Marshal},\text{Bomb}\}$.\\
262 & Whether the piece in the square belongs to the player, and has ever protected an empty square.\\
263 & Whether the piece in the square belongs to the player, and has ever protected a piece of the player which, at the time of the protection, was not revealed.\\
\end{channeltable}
\begin{channeltable}
264--274 & Whether the piece in the square belongs to the player, and has ever protected against a piece of the opponent which, at the time of the protection, had revealed type $T \in \{\text{Spy},\dots,\text{Marshal},\text{Bomb}\}$.\\
275 & Padding channel (all zeros).\\
276 & Whether the piece in the square belongs to the player, and has ever protected against a piece of the opponent which, at the time of the protection, was not revealed.\\
\end{channeltable}
\begin{channeltable}
277--287 & Whether the piece in the square belongs to the player, and has ever been protected by a piece of the player which, at the time of the protection, had revealed type $T \in \{\text{Spy},\dots,\text{Marshal},\text{Bomb}\}$.\\
288 & Padding channel (all zeros).\\
289 & Whether the piece in the square belongs to the player, and has ever been protected by a piece of the player which, at the time of the protection, was not revealed.\\
\end{channeltable}
\begin{channeltable}
290--300 & Whether the piece in the square belongs to the player, and has ever been protected against a piece of the opponent which, at the time of the protection, had revealed type $T \in \{\text{Spy},\dots,\text{Marshal},\text{Bomb}\}$.\\
301 & Padding channel (all zeros).\\
302 & Whether the piece in the square belongs to the player, and has ever been protected against a piece of the opponent which, at the time of the protection, was not revealed.\\
\end{channeltable}
\begin{channeltable}
303--354 & Same as channels 251--302, but for the opponent.\\
\end{channeltable}

\paragraph{Additional information}

In addition, the network has access to the following information.

\begin{channeltable}
355--454 & Starting location of each piece on the board, one-hot encoded. Specifically, channel $355+k$ encodes in position $(i,j) \in \{1,\dots, 10\}^2$ whether the piece currently occupying square $(i,j)$ started from square $k$, according to a row-major ordering of the squares.\\
\end{channeltable}

\subsection{Representation of Recent Moves}

In addition to the board state channels described above, we also include channels that encode the most recent 32 moves. Each move is encoded in a channel, for a total of 32 additional channels. For each move, a $10\times 10$ matrix of floating point numbers encodes the source and destination square of the movement, with the convention that a $+1.0$ is written in the destination square and a $-1.0$ in the source square. All other squares are filled with zeros.

\section{Stratego Implementation Details}
\subsection{Game Rules}
\label{app:game rules}

There are two main parameterized game settings, detailed below. One is the value of $k$ for the $k$-move rule, which states that a draw is declared after $k$ consecutive battleless moves. For training, we used a 100-move rule. We found that this outperformed training under a 200-move rule---even though evaluation took place under a 200-move rule. (The representation of the proximity of the rule to triggering is fractional---see channel 42 in Section~\ref{app:board-state}; so $n$ moves without a battle was represented as $n$/100 during training, but $n$/200 during testing.) This may be because it was effective at discouraging dawdling, thereby increasing the quality of the generated data. The other game setting is the maximum game length before a draw is declared, which was set to $4{,}000$. This was imposed for edge-case safety and is not an actual rule of Stratego. In practice, we found that it virtually never triggered when paired with a 100-move rule.

\begin{detailtable}{Game rules training hyperparameters.}{tab:game-hyper}{Hyperparameter}{Value}
Number of consecutive battleless moves after which a draw is declared & $100$\\
Number of moves after which a draw is declared & $4{,}000$\\
\end{detailtable}

\subsection{Data Collection}

For data collection, each GPU simulated $1{,}536$ independent environments in parallel, playing out 202 moves (101 for each player) for each of these environments between training iterations. (Over 16 GPUs, this yielded about 5 million transitions per training iteration.) The initial boards of new games were sampled using a pool of pre-generated setups (sampled from the setup network), which contained $1{,}000$ setups per player per GPU and was regenerated after each training iteration. These hyperparameters are listed below.

\begin{detailtable}{Data collection hyperparameters.}{tab:data-hyper}{Hyperparameter}{Value}
Number of parallel environments & $1{,}536$ per GPU\\
Generated setups per player & $1{,}000$ per GPU\\
Number of moves between training iterations & $202$\\
\end{detailtable}

\subsection{Setup Learning} \label{app:setup}

Given a setup prefix (meaning a setup partially filled in under row-major order), the setup network produces three outputs:
\begin{enumerate}
    \item A categorical estimate of the probabilities of the possible outcomes of the game (win, loss, draw), given the setup prefix.
    \item A real-valued estimate of the conditional entropy of the setup, given the setup prefix.
    \item A distribution over the next piece placement, given the setup prefix.
\end{enumerate}
We trained these outputs on all setups for games that finished during the previous data collection period. Since most games span multiple reinforcement learning iterations, the training data was slightly off-policy.

\begin{detailtable}{Setup learning notation.}{tab:setup-learning-notation}{Symbol}{Meaning}
$\bar{\sigma}$ & Setup for which a game was completed.\\
$\sigma$ & A prefix of $\bar{\sigma}$.\\
$\theta_t$ & Parameters that generated $\bar{\sigma}$.\\
$\theta$ & Current parameters.\\
$\pi_{\theta_t}(\bar{\sigma} \mid \sigma)$ & Probability of $\bar{\sigma}$ given $\sigma$ under $\theta_t$.\\
$h_{\theta_t}(\sigma)$ & Predicted conditional entropy of setup prefix $\sigma$ under $\theta_t$.\\
$h_{\theta}(\sigma)$ & Predicted conditional entropy of setup prefix $\sigma$ under $\theta$.\\
$o$ & Outcome in $\{\text{win=1, loss=-1, draw=0}\}$ of the game started from $\bar{\sigma}$.\\
$v_{\theta}(o \mid \sigma)$ & Predicted probability of outcome $o$ given $\sigma$ under $\theta$.\\
$\mathbb{E}[v_{\theta_t}(\sigma)]$ & Predicted expected outcome (with win=1, loss=-1, draw=0) given $\sigma$ under $\theta_t$.\\
$\alpha$ & Regularization temperature (coefficient for entropy maximization).\\
$\sigma^{+}$ & Immediate successor to $\sigma$ on the path to $\bar{\sigma}$.\\
$\pi_{\theta_t}(\sigma^{+}\mid \sigma)$ & Probability of $\sigma^{+}$ given $\sigma$ under $\theta_t$.\\
$\pi_{\theta}(\sigma^{+}\mid \sigma)$ & Probability of $\sigma^{+}$ given $\sigma$ under $\theta$.\\
$\pi_{\theta_t}(\sigma)$ & Distribution over next piece type given $\sigma$ under $\theta_t$.\\
$\pi_{\theta}(\sigma)$ & Distribution over next piece type given $\sigma$ under $\theta$.\\
\end{detailtable}

With the notation above, we used the conditional entropy prediction loss
\begin{align}
\mathcal{L}_{h}
&= \left(\frac{- \log \pi_{\theta_t}(\bar{\sigma} \mid \sigma)}{10} - h_{\theta}(\sigma)\right)^2;
\label{eq:conditional_entropy_pred}
\end{align}
the value prediction loss
\begin{align}
\mathcal{L}_{v}
&= - \log v_{\theta}(o \mid \sigma);
\label{eq:setup_value_loss}
\end{align}
the estimated advantage
\[\delta = (o - \mathbb{E}[v_{\theta_t}(\sigma)]) + \alpha (-\log \pi_{\theta_t}(\bar{\sigma} \mid \sigma) -  10 h_{\theta_t}(\sigma));\]
and the policy loss
\begin{align}
\mathcal{L}_{\pi} = -\min\left( r \delta, \text{clip}(r, 0.8, 1.2) \delta \right) + 0.1 \text{KL}(\pi_{\theta}(\sigma), \pi_{\theta_t}(\sigma)),
\label{eq:setup_policy_loss}
\end{align}
where
\[r= \frac{\pi_{\theta}(\sigma^{+} \mid \sigma)}{\pi_{\theta_t}(\sigma^{+} \mid \sigma)}.\]
We weighted these losses as shown below:
\begin{align}
\mathcal{L}_{\text{setup}} = \mathcal{L}_{\pi} + \frac{1}{2} \mathcal{L}_v + \mathcal{L}_h.
\end{align}
We clipped the gradient norm for this loss at $0.5$. We performed $5$ epochs over the setups associated with games that finished during the previous data collection period in batches of $1{,}024$, using Adam with a learning rate of $5 \times 10^{-5}$. We updated an exponential moving average of the parameters---which was used for evaluation---after completion of the training iteration with a smoothing parameter of $0.999$.

These hyperparameter choices are summarized below.
\begin{detailtable}{Setup learning hyperparameters.}{tab:setup-learning-hyper}{Hyperparameter}{Value}
Adam learning rate & $5 \times 10^{-5}$\\
Batch size & $1{,}024$ per GPU\\
Number of epochs per training iteration & $5$\\
Importance ratio clipping parameter & $0.2$\\
Conditional entropy prediction loss coefficient & $1$\\
Reverse KL to data collection policy loss coefficient & $0.1$\\
Value loss coefficient & $0.5$\\
Regularization temperature & $\displaystyle \frac{0.1}{(\text{Training iteration number})^{0.3}}$\\
Maximum gradient norm & $0.5$\\
Normalizing constant for conditional entropy prediction & $1/10$\\
Exponential moving average smoothing factor & $0.999$\\
\end{detailtable}

\subsection{Move Learning} \label{app:move}

Given a position, the move network produces two outputs:
\begin{enumerate}
    \item A categorical estimate of the probabilities of the possible outcomes of the game (win, loss, draw), given the position.
    \item A distribution over moves that are legal given the position.
\end{enumerate}
We trained these outputs on some but not all of the moves made over the last 202 simulator steps. Specifically, we included a move in the training data if both of the following were true:
\begin{enumerate}
\item The magnitude of the estimated advantage was at or above the 0.75 quantile across the data for which advantages were estimated.
\item The magnitude of the estimated advantage was 0.01 or above.
\end{enumerate}

\begin{detailtable}{Move learning notation.}{tab:move-learning-notation}{Symbol}{Meaning}
$x$ & A position of the acting player, represented as described in Section~\ref{app:infostate}.\\
$\{x'\}$ & The sequence of positions subsequent to $x$ for the same player.\\
$\theta_t$ & The parameters that played the move at position $x$.\\
$\theta$ & The current parameters.\\
$v_{\theta_t}(x)$ & Predicted win, loss, draw probabilities at $x$ under $\theta_t$.\\
$v_{\theta}(x)$ & Predicted win, loss, draw probabilities at $x$ under $\theta$.\\
$\mathbb{E}[v_{\theta_t}(x)]$ & Predicted expected outcome (with win=1, loss=-1, draw=0) at $x$ under $\theta_t$.\\
$o$ & The outcome of the game in which $x$ occurred (if game has finished).\\
$\delta$ & $\lambda$-return-estimated advantage for move played at $x$.\\
$\xi$ & $\lambda$-return-estimated probability of winning, losing, and drawing from $x$.\\
$m$ & The move that the acting player played at $x$.\\
$\pi_{\theta_t}(m \mid x)$ & The probability of $m$ at $x$ under $\theta_t$.\\
$\pi_{\theta}(m \mid x)$ & The probability of $m$ at $x$ under $\theta$.\\
$\rho$ & Magnet policy that selects piece to move uniformly and moves it uniformly.\\
$\alpha$ & Reverse KL to magnet policy $\rho$ loss coefficient.\\
$\pi_{\theta}(x)$ & Probability of each move at $x$ under $\theta$.\\
$\pi_{\theta_t}(x)$ & Probability of each move at $x$ under $\theta_t$.\\
$\rho(x)$ & Probability of each move at $x$ under $\rho$.\\
\end{detailtable}

Using the notation above, we estimated the advantage $\delta$ using the $\lambda$-return with $\lambda=0.5$ over the values $\{\mathbb{E}[v_{\theta_t}(x')]\}$ (and $o$ if the game finished) and baseline $\mathbb{E}[v_{\theta_t}(x)]$. We estimated outcome probabilities $\xi$ using the $\lambda$-return with $\lambda=0.8$ over the vector values $\{v_{\theta_t}(x')\}$ (and $o$, as a one-hot vector, if the game finished).

Given these estimates, we updated the move network using the value loss
\begin{align}
\mathcal{L}_v = \text{cross-entropy}(\xi, v_{\theta}(x)),
\label{eq:move-value-loss}
\end{align}
and policy loss
\begin{align}
\mathcal{L}_{\pi} = -\min\left(r \delta, \text{clip}(r, 0.8, 1.2) \delta \right) + 0.1 \text{KL}(\pi_{\theta}(x), \pi_{\theta_t}(x)) + \alpha \text{KL}(\pi_\theta(x), \rho(x)),
\label{eq:move-policy-loss}
\end{align}
where
\[r = \frac{\pi_{\theta}(m \mid x)}{\pi_{\theta_t}(m \mid x)}.\]
We weighted these losses evenly, as shown below:
\[\mathcal{L}_{\text{move}} = \mathcal{L}_{\pi} + \mathcal{L}_v.\]

We trained on this loss in 202 batches of positions grouped by simulator step. If not for filtering by advantage magnitude, the size of these batches would have been $1{,}536$ per GPU; with advantage filtering, they were 1/4 or less of that size. We updated the network using Adam with the learning rate schedule specified below and a maximum gradient norm of $0.267$. We performed only one epoch over this training data. Thereafter, we updated an exponential moving average of the parameters---which was used for evaluation---with a smoothing parameter of 0.999.

\begin{detailtable}{Move learning hyperparameters.}{tab:move-learning-hyper}{Hyperparameter}{Value}
Importance ratio clipping parameter & $0.2$\\
Advantage filtering quantile threshold & $0.75$\\
Advantage filtering magnitude threshold & $0.01$\\
Exponential moving average smoothing factor & $0.999$\\
Advantage estimation $\lambda$ & $0.5$\\
Outcome estimation $\lambda$ & $0.8$\\
Reverse KL to data collection policy loss coefficient & $0.1$\\
Adam learning rate & $\displaystyle \text{clip}\left(\frac{0.5}{(\text{Training iteration number})^{1.1}}, 5 \times 10^{-6}, 1 \times 10^{-4} \right)$\\
Reverse KL to magnet policy loss coefficient & $\displaystyle \frac{0.05}{(\text{Training iteration number})^{0.3}}$\\
Maximum gradient norm & $0.267$\\
Number of epochs per training iteration & $1$\\
Value loss coefficient & $1$\\
Batch size before filtering & $1{,}536$ per GPU\\
\end{detailtable}

\subsection{Belief Learning} \label{app:belief}

Given a position, the belief network produces a distribution over configurations of opponent hidden pieces. We trained the belief network on every position from a stream of self-play games of the final setup and move networks, with no filtering or subsampling. At each position, the input was the associated information state---represented as described in Section~\ref{app:infostate}---and the target was the ground-truth types of the opponent hidden pieces. Training used teacher forcing to minimize the negative log likelihood of these true opponent hidden piece types.

The associated training hyperparameters are summarized below.
\begin{detailtable}{Belief learning hyperparameters.}{tab:belief-learning-hyper}{Hyperparameter}{Value}
Adam learning rate & $5 \times 10^{-5}$\\
Maximum gradient norm & $0.5$\\
Batch size & game length$/2$\\
Data type & bfloat16\\
Exponential moving average smoothing factor & $0.99$\\
Number of environments & $1{,}024$\\
\end{detailtable}

\subsection{Networks}

Details about the networks are given below. Pre-layernorm (i.e., layernorm~\cite{layernorm} before the sublayer rather than after the residual~\cite{resnet}) was used for all of the networks. The belief architecture is detailed in Extended Data Fig.~\hyperref[fig:belief-arch]{4}.

\begin{detailtable}{Setup network information.}{tab:setup-network-hyper}{Hyperparameter}{Value}
Depth & 4\\
Embedding dimension & 512\\
Number of heads & 8\\
Learned positional embedding initialization standard deviation & $0.1$\\
Feedforward dimension & $2{,}048$\\
Total parameters & $12.6$ million\\
\end{detailtable}
\begin{detailtable}{Move network information.}{tab:move-network-hyper}{Hyperparameter}{Value}
Depth & $8$\\
Embedding dimension & $384$\\
Number of heads & $8$\\
Learned positional embedding initialization standard deviation & $0.1$\\
Feedforward dimension & $1{,}536$\\
Total parameters & $14.7$ million\\
\end{detailtable}
\begin{detailtable}{Belief network information.}{tab:belief-network-hyper}{Hyperparameter}{Value}
Encoder depth & $6$\\
Number of decoder blocks & $4$\\
Number of heads & $8$\\
Embedding dimension & $512$\\
Dropout & $0.2$\\
Learned positional embedding initialization & Kaiming uniform~\cite{kaiming_uniform}\\
Feedforward dimension & $2{,}048$\\
Total parameters & $57.1$ million\\
\end{detailtable}
\subsection{Search}

The search proceeds in the following steps.
\begin{enumerate}
    \item The search samples roughly \[\frac{1{,}000}{\text{number of legal actions}}\] possible configurations of the types of the hidden pieces of the opponent from the belief network, given the position $x$.
    \item The search performs $1{,}000$ rollouts of depth 40. There is at least one rollout for each combination of legal move and sampled opponent configuration. After the initial move, which is forced, the subsequent 39 moves are played according to the move network.
    \item The search computes average values $\hat{q}$ for each legal move according to the value output of the move network at the positions reached by the rollouts starting from that move.
    \item The search plays
    \[M \sim \pi_{\text{search}}\]
    where
    \begin{align}
    \pi_{\text{search}} &= \operatorname*{arg\,max}_{\pi} \, \langle \hat{q}, \pi \rangle - \alpha \text{KL}(\pi, \rho) - \beta \text{KL}(\pi, \pi_{\theta})\\
    &\propto \left[e^{\hat{q}} \rho^{\alpha} \pi_{\theta}^{\beta} \right]^{\frac{1}{\alpha + \beta}}
    \end{align}
    and where $\rho = \rho(x)$ is the magnet policy distribution over moves at position $x$, $\pi_{\theta} = \pi_{\theta}(x)$ is the move network distribution over moves at position $x$, $\alpha$ is the reverse KL coefficient to the magnet policy, and $\beta$ is the reverse KL coefficient to the move network policy.
\end{enumerate}
The hyperparameters for this procedure are summarized below.

\begin{detailtable}{Search hyperparameters.}{tab:search-hyper}{Hyperparameter}{Value}
Reverse KL to magnet policy loss coefficient & $0.002$\\
Reverse KL to policy network loss coefficient & $0.02$\\
Number of rollouts & $1{,}000$\\
Rollout depth & $40$\\
\end{detailtable}

\section{Stratego Training Logs}
\begin{figure}[!htbp]
    \centering
    \includegraphics[width=.47\linewidth,page=1]{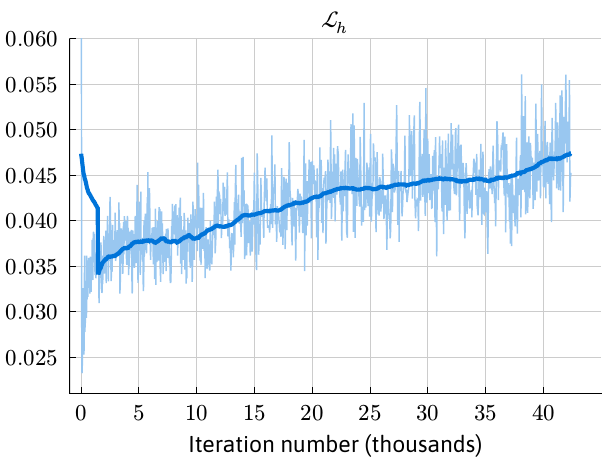}\hfill
    \includegraphics[width=.47\linewidth,page=2]{figures/setup_figures.pdf}\\[2ex]
    \includegraphics[width=.47\linewidth,page=3]{figures/setup_figures.pdf}\hfill
    \includegraphics[width=.47\textwidth,page=4]{figures/setup_figures.pdf}
    \caption{Loss values for setup learning over the training run.} \label{fig:setup-training-logs}
\end{figure}

\begin{figure}[!htbp]
    \centering
    \includegraphics[width=.47\textwidth,page=5]{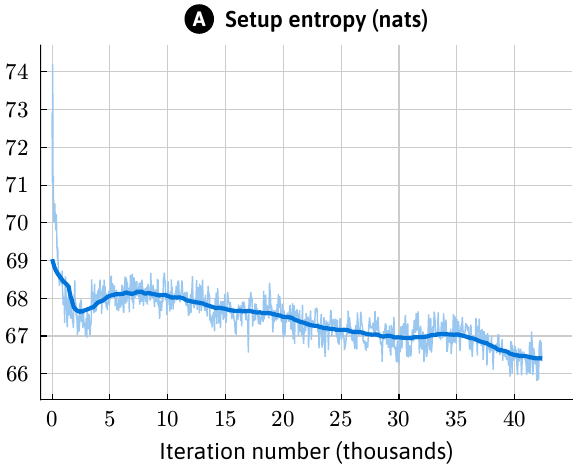}\hfill
    \includegraphics[width=.47\textwidth,page=6]{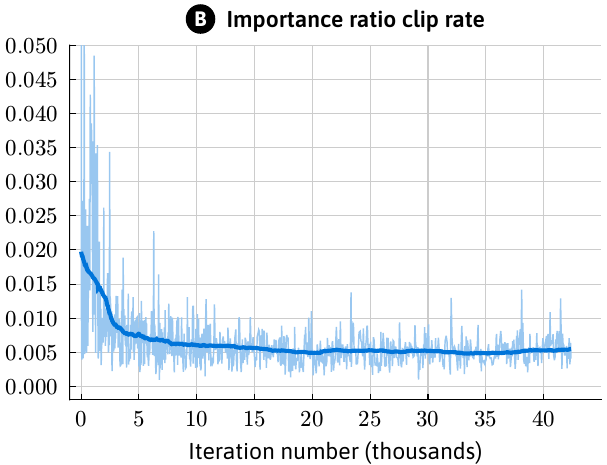}\\[2ex]
    \includegraphics[width=.47\textwidth,page=7]{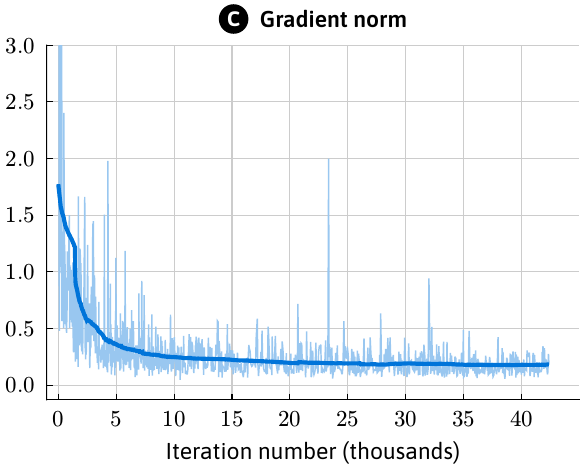}\hfill
    \includegraphics[width=.47\textwidth,page=8]{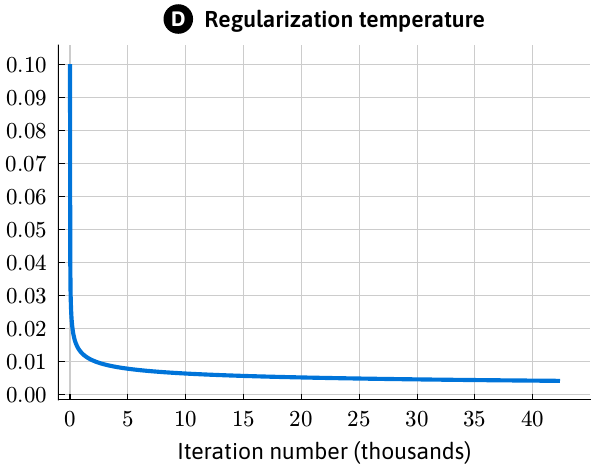}
    \caption{Other setup-related values over the training run. \textbf{a} Entropy of the setup distribution.
    \textbf{b} Proportion of data whose importance ratio is clipped.
    \textbf{c} Norm of the gradient of the loss.
    \textbf{d} Entropy maximization coefficient $\alpha$.} \label{fig:setup-other-logs}
\end{figure}

\begin{figure}[!htbp]
  \centering
  \includegraphics[width=.48\textwidth,page=1]{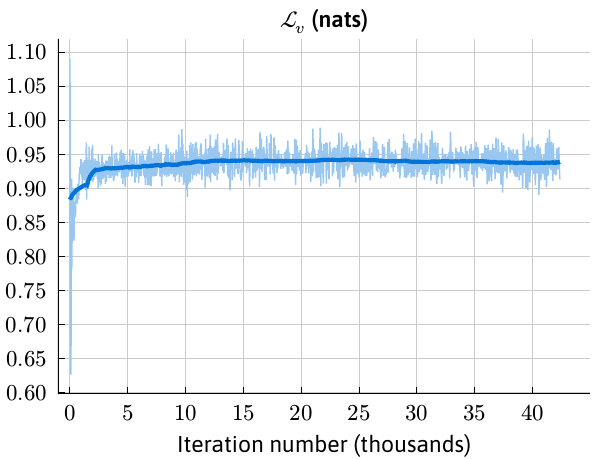}\hfill\includegraphics[width=.48\textwidth,page=2]{figures/move_plots.pdf}\\[2ex]
  \includegraphics[width=.48\textwidth,page=3]{figures/move_plots.pdf}\hfill\includegraphics[width=.48\textwidth,page=4]{figures/move_plots.pdf}\caption{Loss values for move learning over the training run.}
  \label{fig:move-training-logs}
\end{figure}

\begin{figure}[!htbp]
  \centering
  \includegraphics[width=.48\textwidth,page=5]{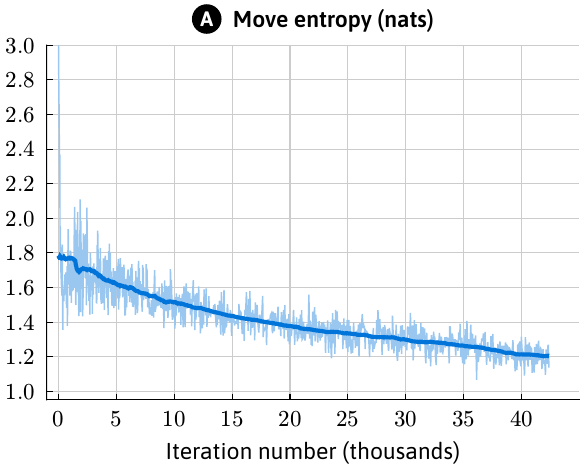}\hfill
  \includegraphics[width=.48\textwidth,page=6]{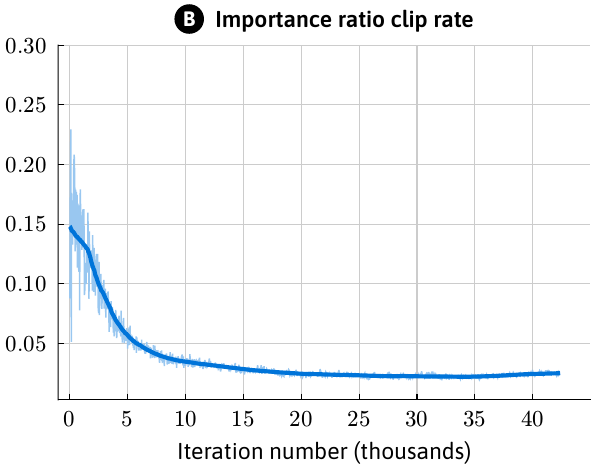}\\[2ex]
  \includegraphics[width=.48\textwidth,page=7]{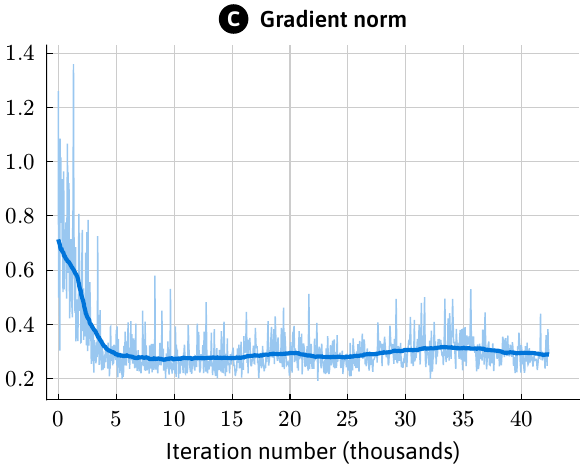}\hfill
  \includegraphics[width=.48\textwidth,page=8]{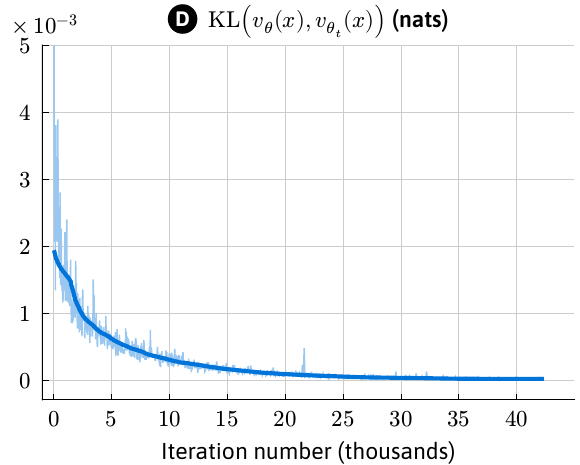}\\[2ex]
  \includegraphics[width=.48\textwidth,page=9]{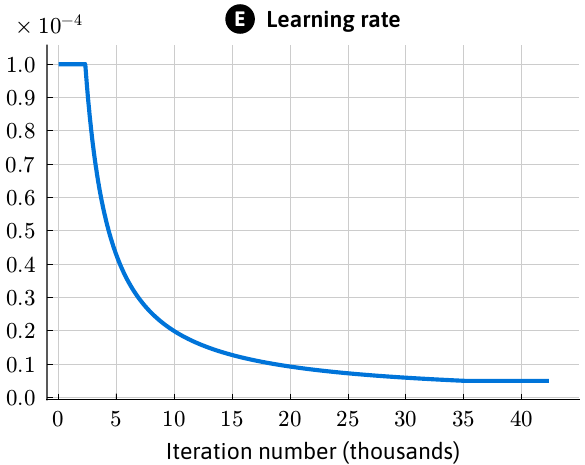}\hfill
  \includegraphics[width=.48\textwidth,page=10]{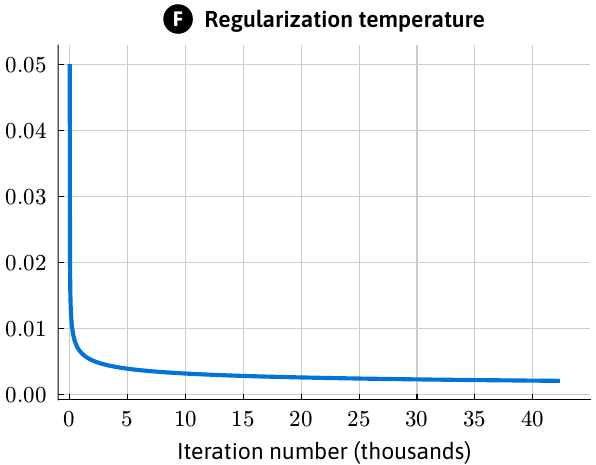}
  \caption{Other move-related values over the training run. \textbf{a} Entropy of moves over self-play distribution. \textbf{b} Proportion of data whose importance ratio is clipped. \textbf{c} Norm of the gradient of the loss. \textbf{d} Value function change. \textbf{e} Adam learning rate. \textbf{f} Coefficient to magnet policy $\alpha$.} \label{fig:move-other-logs}
\end{figure}

We show loss values and other quantities for the setup network over the training run in Figure~\ref{fig:setup-training-logs} and Figure~\ref{fig:setup-other-logs}. We show loss values and other quantities for the move network over the training run in Figure~\ref{fig:move-training-logs} and Figure~\ref{fig:move-other-logs}. For both networks, dynamic damping guides the policy from taking large steps in a heavily regularized regime at the start of training to taking small steps in a weakly regularized regime by the end of training. For the setup network, this can be observed by the simultaneous reduction of 1) the entropy of the network and 2) both the KL divergence relative to the network that generated the setup and the importance ratio clip rate. For the move network, this can be observed by the simultaneous 1) increase in KL divergence relative to the magnet policy and decrease in entropy and 2) decrease in both the KL divergence relative to the policy that made the move and the importance ratio clip rate.

\begin{figure}[!htbp]
  \centering
  \includegraphics[height=5.1cm]{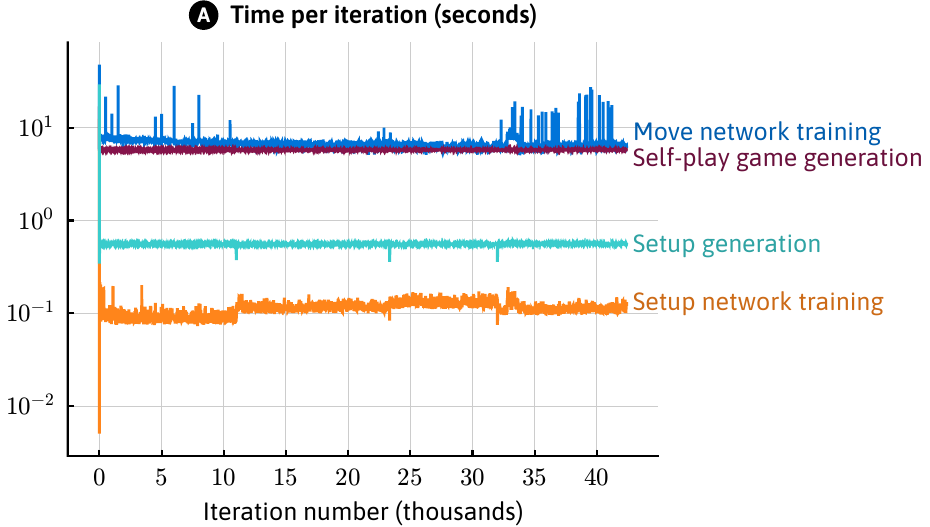}\includegraphics[height=5.1cm]{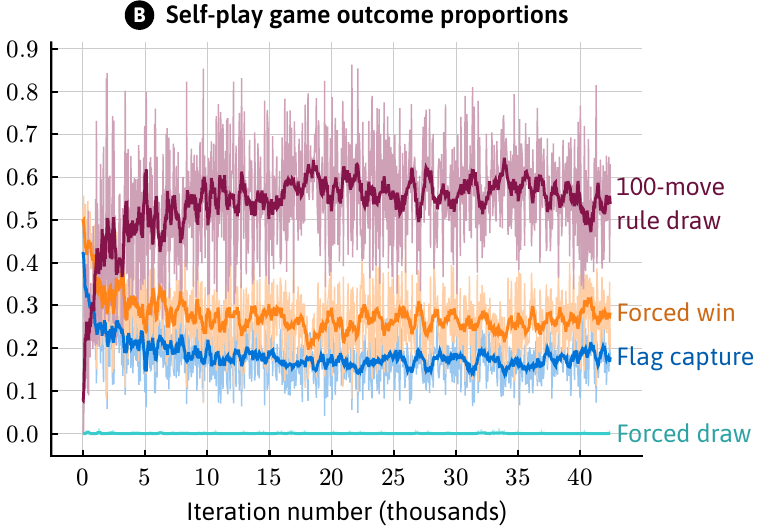}\\[2ex]
  \includegraphics[height=5.1cm,page=1]{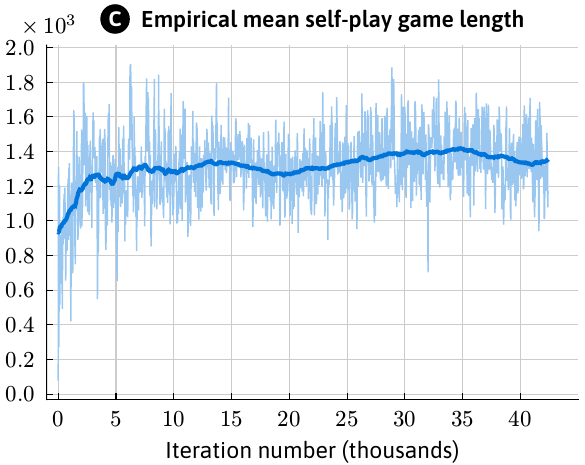}\hskip2cm
  \includegraphics[height=5.1cm,page=2]{figures/expected_game_prop.pdf}\\[2ex]
  \caption{Other information about the training process and games. \textbf{a} Time expenditure of different components of the reinforcement learning loop. Iteration speed is bottlenecked by training and data collection for the move network. \textbf{b} Proportion over training of each of the four possible game endings: Flag capture, forced win (i.e., by virtue of the other player having no legal moves), forced draw (i.e., by virtue of neither player having legal moves), and a 100-move rule draw (i.e., by virtue of there having been 100 consecutive battleless moves). Ataraxos playing against itself draws a much higher proportion of games than humans playing against one another. \textbf{c} The average length of self-play games over training. Ataraxos playing against itself results in much longer games than humans playing against one another. \textbf{d} The average return (win=1, loss=-1, draw=0) for the red player (the first moving player) over training. The roughly equal returns of the red and blue players in the self-play of Ataraxos games support the human belief that move priority is not strategically significant in Stratego.} \label{fig:other-logs}
\end{figure}

Figure~\ref{fig:other-logs} shows other information about the training process and games, including training times, game outcomes, game lengths, and move priority significance.

\begin{figure}[!htbp]
    \centering
    \includegraphics[width=.94\linewidth]{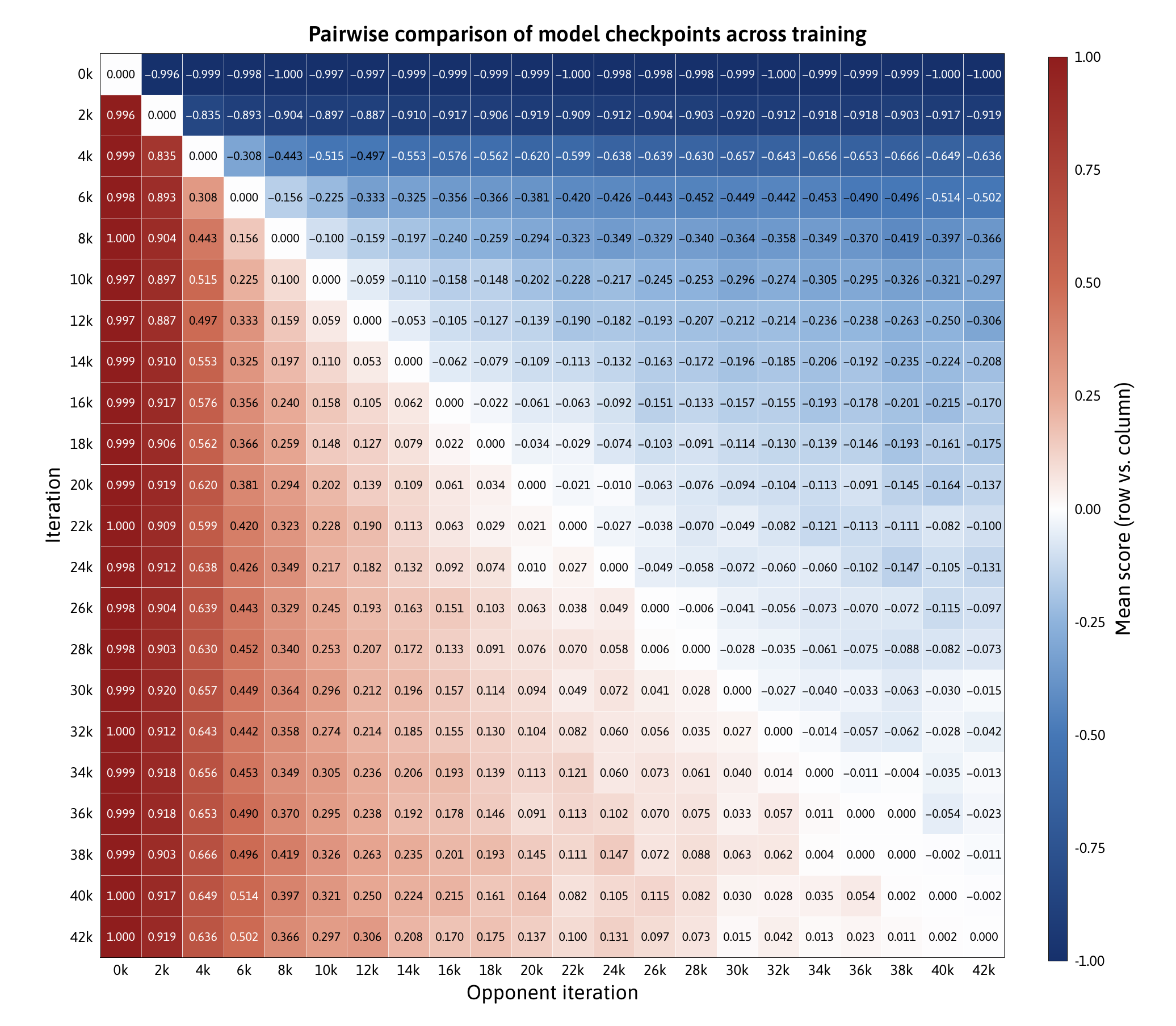}
    \caption{Pairwise evaluation of 22 checkpoints from the training run (those whose iteration indices were divisible by $2{,}000$). Raw checkpoint pairs were evaluated over $2{,}000$ games. The $32{,}000$-iteration row and column average two evaluation blocks and therefore aggregate $4{,}000$ games per off-diagonal entry; all other off-diagonal entries use $2{,}000$ games. Each entry shows the average score of the row checkpoint against the column checkpoint, with wins scored as $+1$, draws as $0$, and losses as $-1$.}
    \label{fig:pairwise}
\end{figure}

\FloatBarrier
To assess training progress more directly, we performed pairwise evaluations of checkpoints from the training run whose iteration indices were divisible by $2{,}000$, playing $2{,}000$ games per raw checkpoint pair. The $32{,}000$-iteration checkpoint occurred in two overlapping checkpoint exports, so we averaged its two evaluation blocks into one displayed row and column; entries involving that checkpoint therefore aggregate $4{,}000$ games, whereas the other off-diagonal entries use $2{,}000$ games. Each entry of the resulting 22-checkpoint score matrix gives the average score of the row checkpoint against the column checkpoint, with wins scored as $+1$, draws as $0$, and losses as $-1$. As shown in Figure~\ref{fig:pairwise}, later checkpoints generally outperform earlier ones, consistent with approximately monotonic improvement over the course of training.

\clearpage

\section{Barrage Stratego Implementation Details}
\label{app:barrage}

Our approach to Barrage Stratego followed the same design as our approach to classic Stratego, with hyperparameters as specified below.
Some notable differences from the classic-Stratego configuration were a smaller move network, $\lambda$ values of zero for advantage and outcome estimation, a constant rather than annealed learning rate for move learning, and a shallower search.
The policy network was trained on 4 H100s over 12 days.
The belief network was trained on 1 H100 over 2 days.

\subsection{Game Rules}

\begin{detailtable}{Barrage Stratego game rules training hyperparameters.}{tab:barrage-game-hyper}{Hyperparameter}{Value}
Number of consecutive battleless moves after which a draw is declared & $100$\\
Number of moves after which a draw is declared & $2{,}000$\\
\end{detailtable}

\subsection{Data Collection}

\begin{detailtable}{Barrage Stratego data collection hyperparameters.}{tab:barrage-data-hyper}{Hyperparameter}{Value}
Number of parallel environments & $1{,}024$ per GPU\\
Generated setups per player & $1{,}000$ per GPU\\
Number of moves between training iterations & $202$\\
\end{detailtable}

\subsection{Setup Learning}

\begin{detailtable}{Barrage Stratego setup learning hyperparameters.}{tab:barrage-setup-learning-hyper}{Hyperparameter}{Value}
Adam learning rate & $5 \times 10^{-5}$\\
Batch size & $1{,}024$ per GPU\\
Number of epochs per training iteration & $2$\\
Importance ratio clipping parameter & $0.2$\\
Conditional entropy prediction loss coefficient & $1$\\
Reverse KL to data collection policy loss coefficient & $0.1$\\
Value loss coefficient & $0.5$\\
Regularization temperature & $\displaystyle \text{clip}\!\left(\frac{0.2}{(\text{Training iteration number})^{0.2}},\ 0.01,\ 1.0\right)$\\
Maximum gradient norm & $0.5$\\
Normalizing constant for conditional entropy prediction & $1/10$\\
Exponential moving average smoothing factor & $0.999$\\
\end{detailtable}

\subsection{Move Learning}

\begin{detailtable}{Barrage Stratego move learning hyperparameters.}{tab:barrage-move-learning-hyper}{Hyperparameter}{Value}
Importance ratio clipping parameter & $0.2$\\
Advantage filtering quantile threshold & $0.5$\\
Advantage filtering magnitude threshold & $0.01$\\
Exponential moving average smoothing factor & $0.999$\\
Advantage estimation $\lambda$ & $0.0$\\
Outcome estimation $\lambda$ & $0.0$\\
Reverse KL to data collection policy loss coefficient & $0.001$\\
Adam learning rate & $2.15 \times 10^{-5}$\\
Reverse KL to magnet policy loss coefficient & $\displaystyle \text{clip}\!\left(\frac{0.04}{(\text{Training iteration number})^{0.3}},\ 0.006,\ 0.1\right)$\\
Maximum gradient norm & $0.4$\\
Number of epochs per training iteration & $2$\\
Value loss coefficient & $0.5$\\
Batch size before filtering & $1{,}024$ per GPU\\
\end{detailtable}

\subsection{Belief Learning}

\begin{detailtable}{Barrage Stratego belief learning hyperparameters.}{tab:barrage-belief-learning-hyper}{Hyperparameter}{Value}
Adam learning rate & $2 \times 10^{-5}$\\
Maximum gradient norm & $0.5$\\
Exponential moving average smoothing factor & $0.99$\\
Number of environments & $1{,}024$\\
\end{detailtable}

\subsection{Networks}

\begin{detailtable}{Barrage Stratego setup network information.}{tab:barrage-setup-network-hyper}{Hyperparameter}{Value}
Depth & $4$\\
Embedding dimension & $512$\\
Number of heads & $8$\\
Learned positional embedding initialization standard deviation & $0.1$\\
Feedforward dimension & $2{,}048$\\
Total parameters & $12.6$ million\\
\end{detailtable}
\begin{detailtable}{Barrage Stratego move network information.}{tab:barrage-move-network-hyper}{Hyperparameter}{Value}
Depth & $7$\\
Embedding dimension & $288$\\
Number of heads & $8$\\
Learned positional embedding initialization standard deviation & $0.1$\\
Feedforward dimension & $864$\\
Total parameters & $6.1$ million\\
\end{detailtable}
\begin{detailtable}{Barrage Stratego belief network information.}{tab:barrage-belief-network-hyper}{Hyperparameter}{Value}
Encoder depth & $6$\\
Decoder depth & $6$\\
Number of heads & $8$\\
Embedding dimension & $512$\\
Feedforward dimension & $2{,}048$\\
Dropout & $0.1$\\
Total parameters & $56.9$ million\\
\end{detailtable}

\subsection{Search}

The search procedure is analogous to that of classic Stratego, with the update parameterized by a step size and temperature rather than by two KL coefficients.

\begin{detailtable}{Barrage Stratego search hyperparameters.}{tab:barrage-search-hyper}{Hyperparameter}{Value}
Number of rollouts & $1{,}024$\\
Rollout depth & $10$\\
Update step size & $10$\\
Magnet policy regularization (step size $\times$ temperature) & $0.01$\\
Minimum samples per legal action & $10$\\
\end{detailtable}

\section{Hanabi Implementation Details}

The policy networks were each trained on a single L40 over 2 days.
The belief networks were each trained on a single L40 over 1 day.

\subsection{Actor Learning}
\label{app:hanabi-actor-learning}

We trained the base policy using dynamically damped self-play reinforcement learning, with the annealing schedules and other hyperparameters listed below.

\begin{detailtable}{Hanabi actor learning hyperparameters (shared across 2--5 players).}{tab:hanabi-actor-learning-hyper}{Hyperparameter}{Value}
Importance ratio clipping parameter & $0.2$\\
Exponential moving average smoothing factor & $0.9999$\\
Advantage estimation $\lambda$ & $0.83$\\
Discount factor $\gamma$ & $0.999$\\
Adam learning rate & $\displaystyle 3.5 \times 10^{-4}\!\left(1 - \left(\frac{t}{T}\right)^{\!0.4}\right)$\\
Entropy coefficient & $\displaystyle 0.035\!\left(1.05 - 0.2\left(\frac{t}{T}\right)^{\!2}\right)$\\
Schedule horizon $T$ & $10^{7}$\\
Maximum gradient norm & $0.5$\\
Value loss coefficient & $0.5$\\
Batch size & $256$\\
Number of parallel environments & $2{,}048$\\
\end{detailtable}

\subsection{Belief Learning}
\label{app:hanabi-belief-learning}

For belief training, we followed the practice and implementation from prior work~\cite{lbs}, with the hyperparameters listed below.

\begin{detailtable}{Hanabi belief learning hyperparameters.}{tab:hanabi-belief-learning-hyper}{Hyperparameter}{Value}
Adam learning rate & $1 \times 10^{-4}$\\
Batch size & $128$\\
Number of training steps & $10^{6}$\\
Number of parallel environments & $4{,}800$\\
\end{detailtable}

\subsection{Networks}
\label{app:hanabi-networks}

We used the exact policy and belief network architectures from prior work~\cite{lbs}, which are summarized below.

\begin{detailtable}{Hanabi actor network architecture.}{tab:hanabi-actor-network}{Hyperparameter}{Value}
Private net & $3$-layer MLP (hidden dim $512$, ReLU)\\
Public net & $1$-layer MLP (hidden dim $512$, ReLU)\\
Public LSTM & $2$-layer LSTM (hidden dim $512$)\\
Policy head & Linear ($512 \to |\mathcal{A}|$)\\
Value head & Linear ($512 \to 1$)\\
Total parameters & $5.9$ million\\
\end{detailtable}
\begin{detailtable}{Hanabi belief network architecture.}{tab:hanabi-belief-network}{Hyperparameter}{Value}
Observation encoder & $2$-layer MLP (hidden dim $512$, ReLU)\\
Sequence & $2$-layer LSTM (hidden dim $512$)\\
Autoregressive card prediction -- Embedding & Linear ($25 \to 64$, no bias)\\
Autoregressive card prediction -- LSTM & $1$-layer LSTM (input dim $576$, hidden dim $512$)\\
Autoregressive card prediction -- Output & Linear ($512 \to 25$)\\
Total parameters & $7.05$--$7.37$ million\\
\end{detailtable}

\subsection{Search}
\label{app:hanabi-search}

For the search, we ran $10{,}000$ simulations at each search step, evenly distributed across all legal actions. Let $A$ denote the set of legal actions. We sampled $\frac{10{,}000}{\lvert A \rvert}$ hands from the belief model and used the same sampled hands for all legal actions during search. We rolled out until the end of the episode to collect cumulative return.

\begin{detailtable}{Hanabi search hyperparameters.}{tab:hanabi-search-hyper}{Hyperparameter}{Value}
Number of rollouts & $10{,}000$\\
Update step size & $30$\\
Rollout temperature & $0$\\
\end{detailtable}

\section{\textit{Dou Dizhu} Implementation Details}

The policy network was trained for 2 days on 4 H100s.
The belief network was trained for 8 hours on 4 H100s.
\subsection{Actor Learning}
\label{sec:actor-learning}

\begin{detailtable}{\textit{Dou Dizhu} actor learning hyperparameters.}{tab:doudizhu-actor-learning-hyper}{Hyperparameter}{Value}
Importance ratio clipping parameter & $\displaystyle0.02 \cdot \min\left\{\frac{5000}{\text{Training iteration number}}, 1\right\}$\\
Exponential moving average smoothing factor & $0.999$\\
Advantage estimation $\lambda$ & $0.95$\\
Outcome estimation $\lambda$ & $0.95$\\
Reverse KL to data collection policy loss coefficient & $0$\\
Muon learning rate & $\displaystyle 4 \times 10^{-4} \cdot \min\left\{\frac{5000}{\text{Training iteration number}}, 1\right\}$\\
Regularization temperature & $\displaystyle 0.1 \cdot \min\left\{\frac{500}{\text{Training iteration number}}, 1\right\}^{0.5}$\\
Maximum gradient norm & $1$\\
Number of epochs per training iteration & $4$\\
Value loss coefficient & $0.5$\\
Number of parallel environments & $512$\\
Trajectories collected per iteration & $2{,}048$\\
Batch size & $512$\\
\end{detailtable}

\subsection{Belief Learning}
\label{sec:belief-learning}

In \textit{dou dizhu}, the hidden state consists of the unseen cards held by the other two players. We therefore trained a belief model that autoregressively predicts card ownership in descending rank order. At each prediction step, the model takes the available observation history and the previously predicted belief tokens as input, and predicts which player holds the next card. The belief model is an autoregressive transformer: each historical game state is encoded by a small MLP and projected to the transformer embedding dimension, while each belief token is represented as the sum of a card embedding and a player embedding. We trained the model from snapshots collected at each time step, using an NLL loss on the belief-token targets. During online search, the model is queried autoregressively to sample or score hidden-card assignments. We visualize the training dynamics in Supplementary Figure~\hyperref[fig:doudizhu-belief-train]{S7}; the hyperparameters are listed below.

\begin{detailtable}{\textit{Dou dizhu} belief learning hyperparameters.}{tab:doudizhu-belief-learning-hyper}{Hyperparameter}{Value}
Muon~\cite{jordan2024muon} learning rate & $4 \times 10^{-4}$\\
WSD scheduler stable ratio & $0.9$\\
Maximum gradient norm & $1.0$\\
Time steps collected per iteration & $2{,}048$\\
\end{detailtable}

\begin{figure}[!htbp]
\centering
\includegraphics[width=\linewidth]{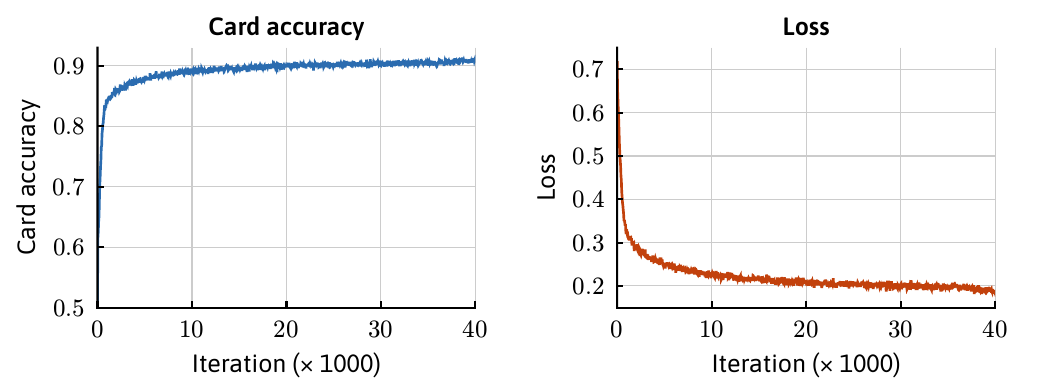}
\caption{\textit{Dou dizhu} belief-model training dynamics. Card-ownership accuracy (left) is the fraction of autoregressive hidden-card predictions for which the model identifies the ground-truth holder of the card. Loss (right) is the corresponding mean negative log-likelihood assigned to the ground-truth holder of the card. Both metrics are averaged over valid hidden-card predictions within each belief sequence and then over sequences.}
\label{fig:doudizhu-belief-train}
\end{figure}
\FloatBarrier

\subsection{Networks}

\begin{detailtable}{\textit{Dou dizhu} actor network information.}{tab:doudizhu-actor-network-hyper}{Hyperparameter}{Value}
Transformer depth & $4$\\
Embedding dimension & $128$\\
Number of heads & $4$\\
Feedforward depth & $3$\\
Feedforward dimension & $256$\\
Total parameters & $1.9$ million\\
\end{detailtable}
\begin{detailtable}{\textit{Dou Dizhu} belief network information.}{tab:doudizhu-belief-network-hyper}{Hyperparameter}{Value}
Transformer depth & $4$\\
Embedding dimension & $256$\\
Number of heads & $4$\\
Feedforward depth & $2$\\
Feedforward dimension & $256$\\
Total parameters & $3.3$ million\\
\end{detailtable}

\subsection{Search}
\label{sec:doudizhu-search}

Due to the large action space in \textit{dou dizhu}, Ataraxos decomposes each action into a sequence of smaller decisions~\cite{vrpo}. Rather than selecting a full action in one step, the policy incrementally generates an action-description prefix: at each step, it chooses one valid option to append to the current prefix. Ataraxos applies search at each of these intermediate decision points, evaluating the valid continuations conditioned on the current prefix. Let $A$ denote the set of valid options at the current step. For each option, the search procedure samples $\lceil 200 / |A| \rceil$ hidden hands from the belief model. It then estimates the value of that option by forcing the option to be taken under each sampled hand and rolling out the policy network until the end of the game to collect the terminal rewards. The final choice is obtained by taking one damped policy-improvement step from the policy network, with update step size $5$.

\begin{detailtable}{\textit{Dou dizhu} search hyperparameters.}{tab:doudizhu-search-hyper}{Hyperparameter}{Value}
Number of rollouts & $200$\\
Update step size & $5$\\
Rollout temperature & $0$\\
\end{detailtable}

\clearpage
\section{Hanabi Score Details}
This section provides the detailed Hanabi evaluation numbers underlying the results reported in the Methods.
\begin{table}[H]
\centering
\caption{Performance of Ataraxos on Hanabi variants with different numbers of players. We evaluate Ataraxos with different numbers of search players in each variant. Entries report means $\pm$ standard errors over $10{,}000$ games per condition.}
\begin{tabular}{lccc}
\toprule
No. of players & No. of search players & Score & Perfect \\
\midrule
\multirow{3}{*}{2}
& 0 & $24.454 \pm 0.008$ & $69.58\% \pm 0.33\%$ \\
& 1 & $24.614 \pm 0.008$ & $75.72\% \pm 0.43\%$ \\
& 2 & $24.654 \pm 0.007$ & $77.53\% \pm 0.42\%$ \\
\midrule
\multirow{4}{*}{3}
& 0 & $24.612 \pm 0.008$ & $77.61\% \pm 0.29\%$ \\
& 1 & $24.780 \pm 0.006$ & $84.77\% \pm 0.36\%$ \\
& 2 & $24.837 \pm 0.005$ & $87.81\% \pm 0.33\%$ \\
& 3 & $24.863 \pm 0.005$ & $89.90\% \pm 0.30\%$ \\
\midrule
\multirow{5}{*}{4}
& 0 & $24.466 \pm 0.008$ & $68.52\% \pm 0.33\%$ \\
& 1 & $24.683 \pm 0.007$ & $77.65\% \pm 0.42\%$ \\
& 2 & $24.776 \pm 0.006$ & $83.50\% \pm 0.37\%$ \\
& 3 & $24.821 \pm 0.005$ & $86.13\% \pm 0.35\%$ \\
& 4 & $24.852 \pm 0.005$ & $88.40\% \pm 0.32\%$ \\
\midrule
\multirow{6}{*}{5}
& 0 & $23.485 \pm 0.012$ & $26.81\% \pm 0.31\%$ \\
& 1 & $24.008 \pm 0.011$ & $39.19\% \pm 0.49\%$ \\
& 2 & $24.213 \pm 0.010$ & $47.43\% \pm 0.50\%$ \\
& 3 & $24.340 \pm 0.009$ & $53.94\% \pm 0.50\%$ \\
& 4 & $24.393 \pm 0.009$ & $56.84\% \pm 0.50\%$ \\
& 5 & $24.410 \pm 0.009$ & $58.05\% \pm 0.49\%$ \\
\bottomrule
\end{tabular}
\label{tab:hanabi-numbers}
\end{table}

\clearpage
\section{\textit{Dou Dizhu} Additional Results}
\begin{figure}[!htbp]
\centering
\includegraphics[width=0.8\linewidth]{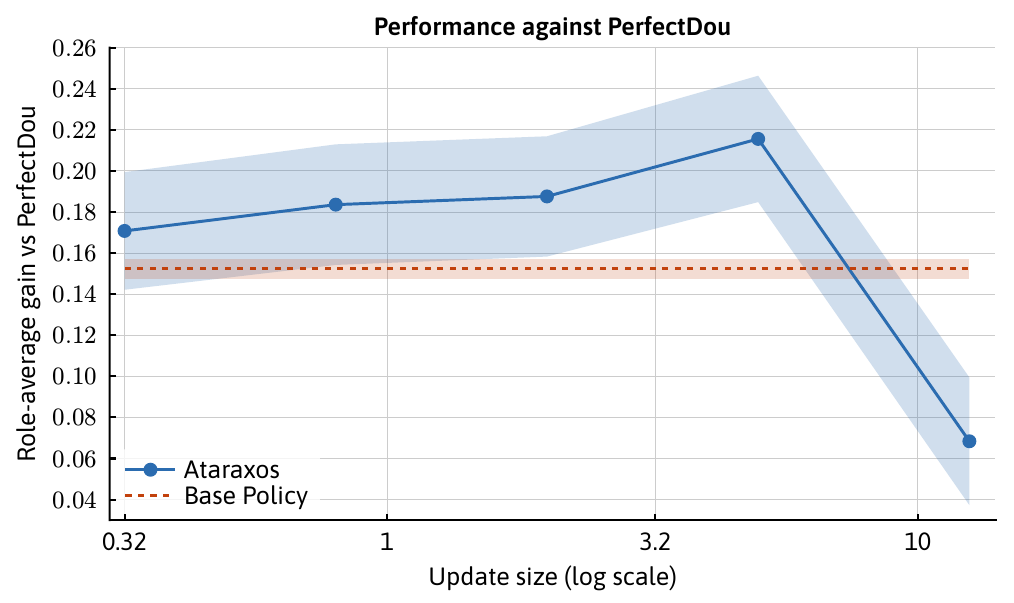}
\caption{Role-averaged score of Ataraxos against PerfectDou under varying search update step sizes. Each setting was evaluated over $2{,}500$ duplicated deals. The dashed line shows the performance of the policy network (i.e., Ataraxos without search).}
\label{fig:doudizhu-update-sizes}
\end{figure}

This section provides the full role-averaged performance matrix (Table~\ref{tab:doudizhu-performance}), the performance breakdown by game role (Table~\ref{tab:doudizhu-performance-by-role}), training diagnostics for the \textit{dou dizhu} belief model (Figure~\ref{fig:doudizhu-belief-train}), and the sweep over update sizes (Figure~\ref{fig:doudizhu-update-sizes}).

\begin{longtblr}[
caption = {\textit{Dou dizhu} role-averaged performance. Entries report means $\pm$ standard errors. Comparisons involving Ataraxos use $10{,}000$ duplicated deals; other comparisons use $100{,}000$ duplicated deals.},
  label = {tab:doudizhu-performance},
  columns = {5},
                ]{hline{2,Z},
                  colspec = {X X X X X},
                  row{even} = {bg=gray!10},
                  row{1} = {font=\bfseries},
                  rowhead = 1,
                 }
Row vs Col & Ataraxos & Policy Network & PerfectDou & DouZero \\
Ataraxos & --- & $0.107 \pm 0.012$ & $0.199 \pm 0.015$ & $0.350 \pm 0.016$ \\
Policy Network & $-0.107 \pm 0.012$ & --- & $0.152 \pm 0.004$ & $0.286 \pm 0.004$ \\
PerfectDou & $-0.199 \pm 0.015$ & $-0.152 \pm 0.004$ & --- & $0.142 \pm 0.004$ \\
DouZero & $-0.350 \pm 0.016$ & $-0.286 \pm 0.004$& $-0.142 \pm 0.004$ & --- \\
\end{longtblr}

The entries in Table~\ref{tab:doudizhu-performance} not involving Ataraxos (i.e., those among the policy network, PerfectDou, and DouZero) are taken from prior evaluations using $100{,}000$ duplicated deals.

We further report performance broken down by role in Table~\ref{tab:doudizhu-performance-by-role}, where the row agent controls the two independent peasants and the column agent plays as the landlord. Comparing Ataraxos with the policy network shows that the search routine provides particularly large gains on the peasant side. This suggests that the search is especially useful for improving coordination between the two cooperating peasant agents.

\begin{longtblr}[
caption = {Performance broken down by game role. The row agent is the peasants; the column agent is the landlord. Entries report means $\pm$ standard errors. Comparisons involving Ataraxos use $10{,}000$ duplicated deals; other comparisons use $100{,}000$ duplicated deals.},
  label = {tab:doudizhu-performance-by-role},
  columns = {5},
                ]{hline{2,Z},
                  colspec = {X X X X X},
                  row{even} = {bg=gray!10},
                  rowhead = 1,
                 }
P \textbackslash \, L & Ataraxos & Policy Network & PerfectDou & DouZero \\
Ataraxos & $0.696 \pm 0.035$ & $0.827 \pm 0.035$ & $0.855 \pm 0.034$ & $1.003 \pm 0.034$ \\
Policy Network & $0.612 \pm 0.035$ & $0.707 \pm 0.011$ & $0.811 \pm 0.011$ & $0.877 \pm 0.011$ \\
PerfectDou & $0.456 \pm 0.036$ & $0.507 \pm 0.011$ & $0.601 \pm 0.011$ & $0.656 \pm 0.011$ \\
DouZero & $0.301 \pm 0.037$ & $0.304 \pm 0.011$& $0.373 \pm 0.011$ & $0.453 \pm 0.011$ \\
\end{longtblr}

\clearpage
\begingroup
\raggedright
\renewcommand{\refname}{Supplementary References}

\endgroup

\end{document}